\documentclass{IEEEoj}
\usepackage{cite}
\usepackage{amsmath,amssymb,amsfonts}
\usepackage{algorithm}
\usepackage{algorithmic}
\usepackage{graphicx,color}
\usepackage{textcomp}
\usepackage{booktabs}
\usepackage{array}
\usepackage{siunitx}
\usepackage{xcolor}

\graphicspath{{figures/}}
\AtBeginDocument{\definecolor{ojcolor}{cmyk}{0.93,0.59,0.15,0.02}}
\def\OJlogo{}
\usepackage{etoolbox}
\makeatletter
\def\arxivnotice{\parbox{\textwidth}{\centering\rffont This work has been submitted to the IEEE for possible publication. Copyright may be transferred without notice, after which this version may no longer be accessible.}}
\def\ps@plain{\def\@oddhead{}\let\@evenhead\@oddhead
  \def\@oddfoot{\vbox to 10pt{\arxivnotice\par\vspace*{2pt}\hbox to \textwidth{\hfill{\rffont\thepage}}}}\let\@evenfoot\@oddfoot}
\def\ps@headings{\def\@oddhead{}%
  \def\@evenhead{\vbox{\hsize\textwidth\vbox to 0pt{\hsize\textwidth\vspace*{7.7pt}\rfxfont\raggedright\rightmark:\ \leftmark\hfill\par}\par\vspace*{16pt}\hbox to \textwidth{\vrule width\textwidth height.3pt depth0pt}}}%
  \def\@evenfoot{\hbox to \textwidth{{\rffont\thepage}\hfill}}\def\@oddfoot{\hbox to \textwidth{\hfill{\rffont\thepage}}}}
\patchcmd{\@maketitle}{\fi.\space\ifx\@editor}{\fi\ifx\@editor}{}{\errmessage{patch failed}}
\patchcmd{\@maketitle}{\ifx\@revised\@empty\else;\ revised\ \@reviseddate\fi}{}{}{\errmessage{patch failed}}
\makeatother

\newtheorem{definitionx}{Definition}
\newtheorem{assumptionx}{Assumption}
\newtheorem{problemx}{Problem}
\newtheorem{propositionx}{Proposition}
\newtheorem{remarkx}{Remark}
\NewDocumentEnvironment{definition}{o}
  {\IfNoValueTF{#1}{\begin{definitionx}}{\begin{definitionx}[#1]}}{\hfill$\square$\end{definitionx}}
\NewDocumentEnvironment{assumption}{o}
  {\IfNoValueTF{#1}{\begin{assumptionx}}{\begin{assumptionx}[#1]}}{\hfill$\square$\end{assumptionx}}
\NewDocumentEnvironment{problem}{o}
  {\IfNoValueTF{#1}{\begin{problemx}}{\begin{problemx}[#1]}}{\hfill$\square$\end{problemx}}
\NewDocumentEnvironment{proposition}{o}
  {\IfNoValueTF{#1}{\begin{propositionx}}{\begin{propositionx}[#1]}}{\end{propositionx}}
\NewDocumentEnvironment{remark}{o}
  {\IfNoValueTF{#1}{\begin{remarkx}}{\begin{remarkx}[#1]}}{\hfill$\square$\end{remarkx}}

\newcommand{\att}{\mathsf{att}}
\newcommand{\ego}{\mathsf{ego}}
\newcommand{\by}{\mathsf{by}}

\newcommand{\oth}{\mathsf{others}}
\newcommand{\Roles}{\mathcal{R}}

\newcommand{\bx}{\mathbf{x}}
\newcommand{\bu}{\mathbf{u}}
\newcommand{\useed}{\mathbf{u}^{\mathrm{seed}}}
\newcommand{\xseed}{\mathbf{x}^{\mathrm{seed}}}
\newcommand{\Uset}{\mathcal{U}}
\newcommand{\Agents}{\mathcal{A}}

\newcommand{\Gop}{\mathbf{G}}
\newcommand{\Fop}{\mathbf{F}}
\newcommand{\Uop}{\mathbf{U}}
\newcommand{\Hop}{\mathbf{H}}
\newcommand{\Oop}{\mathbf{O}}
\newcommand{\own}{\mathrm{own}}
\newcommand{\reads}{\mathrm{reads}}
\newcommand{\kind}{\mathrm{kind}}
\newcommand{\sg}{\mathrm{sg}}
\newcommand{\psiobs}{\psi_{\mathrm{obs}}}
\newcommand{\trel}{t_{\mathrm{rel}}}
\newcommand{\rt}{\tilde\rho}

\newcommand{\pred}[1]{\mathtt{#1}}
\newcommand{\cl}[1]{\texttt{#1}}

\newsavebox{\fitbox}

\makeatletter
\def\subsubsection{\@startsection{subsubsection}{3}{\z@}{-10pt \@plus -2pt \@minus -2pt}{1pt}{\subsubsectionfont}}

\makeatother

\begin{document}
\receiveddate{}
\reviseddate{}
\accepteddate{}
\publisheddate{}
\currentdate{}
\doiinfo{}

\title{Generating Safety-Critical Driving Scenarios from Role-Typed Signal Temporal Logic Specifications Drafted by Large Language Models}

\author{ZIFAN ZENG\affilmark{1,2}, QUNLI ZHANG\affilmark{1}, PENG WANG\affilmark{3}, SHIMING LIU\affilmark{3},
FENG LIU\affilmark{1}, AND ZHENG HU\affilmark{1}}
\affil{RAMS Lab, Huawei Heisenberg Research Center, 80992 Munich, Germany}
\affil{Department of Computer Engineering, Technical University of Munich, 85748 Garching, Germany}
\affil{RAMS Lab, Huawei Technologies Co., Ltd., 518129 Shenzhen, China}
\corresp{CORRESPONDING AUTHOR: Zifan Zeng (e-mail: zifan.zeng@tum.de).}
\markboth{Role-Typed STL Specifications for Scenario Generation}{Zeng \emph{et al.}}

\begin{abstract}
Safety-critical scenarios are rare in recorded driving and must be generated to test and train motion planners of automated vehicles. In guided diffusion models, each scenario type needs its own weighted objective, separate from the validity check. Optimized together with the attacker, the vehicle under test can cancel the attack or draw it toward itself. We state each scenario as a
signal temporal logic specification with clauses typed by role. Every clause names the agent it may
change and the agents it may read, and the vehicle under test reads the attacker only from the attack
onset. A large language model drafts the specification from a description, and a type checker accepts it. A compiler derives the optimization: it edits a diffusion model's samples by the smallest change that satisfies the specification, each clause moves only its owner, the
attacker is solved against the unreacted vehicle under test before its response, and the monitor of the
same specification decides acceptance.
On nuScenes and CommonRoad, judged independently of the
specification, one parameter set produced valid scenarios at a rate 0.14 above a weighted objective tuned per
scenario type, and descriptions yielded at least one valid scenario for 11 of the 16 pre-crash types expressible in the
public data. Learned planners fail more often on the generated scenarios than on their seeds. Fine-tuning on them lowers a planner's failure rate on critical scenarios, and the time headway one clause of the label
requires, not the generator, sets the size of the gain.
\end{abstract}


\maketitle

\section{INTRODUCTION}\label{sec:intro}

\IEEEPARstart{M}{otion} planners of automated vehicles have to be tested in safety-critical
interactions, and learned planners need such interactions in their training data. An example is a
vehicle that cuts in at a short gap in front of the vehicle under test, the ego vehicle. Such interactions
are rare in recorded driving data, and scenarios that contain them therefore have to be
generated~\cite{ding2023survey}.
At the trajectory level, a scenario is the motion of all agents on a map. The content of a scenario is
specified and checked at this level, and the levels below take it as input. Traffic layouts and agent
trajectories condition the generation of driving video and sensor
data~\cite{gao2026scenariocontrol,zhou2025nexus}, and end-to-end driving policies are tested in closed
loop on scenarios defined at this level~\cite{scenepilot2026}.
Trajectory diffusion models trained on recorded data generate realistic multi-agent
behavior~\cite{zhong2023ctg,jiang2023motiondiffuser,scenediffuser2024}, and guiding their sampling
with differentiable objectives makes them controllable~\cite{zhong2023ctg,chang2024safesim,lin2025ccdiff}.
In both uses, the generated trajectory of the ego vehicle has a role of its own. In testing, it shows
that the attack can be avoided, so that a failure of the planner under test is meaningful. In
training, it is the label that the planner imitates.

A generator of such scenarios is useful at scale if it has three capabilities. It has to generalize, that is, cover the scenario types that matter, such as the pre-crash types of the National Highway
Traffic Safety Administration (NHTSA)~\cite{nhtsa2007typology}, at a low cost for each new type. Its
scenarios have to be valid: the attack takes place as specified, and the ego vehicle has a legitimate
response once the attack is observable. Finally, it has to be efficient and yield many valid scenarios per compute hour and per recorded input, since processing a recorded input
further, for example by sensor-level reconstruction, is expensive. This paper addresses scenarios at
the trajectory level. It does not address sensor simulation or the training of the generative model.

A recent survey of foundation models for scenario generation~\cite{foundationmodels2025survey}
lists problems that limit these capabilities. Gradient-guided diffusion models require carefully tuned
objective weights, and adapting them to new constraints is costly. The objectives are hand-written
terms~\cite{zhong2023ctg} or loss code, weights included, that a large language model (LLM) writes from a description~\cite{zhong2023ctgpp,ldscene2025}. Either way, each new scenario type needs its own terms.
Since the objective is separate from the check that later decides validity, a sample can minimize the
objective and still fail the check, and no output carries a mark of validity. The survey also lists the
computational cost of generation among the open problems. Formal specifications are the established way
to state the content of a test scenario, and scenarios are synthesized from
them~\cite{klischat2020synthesizing,finkeldei2023synthesizing}. Signal temporal logic (STL) offers
both a guidance objective, through its quantitative semantics, and a
monitor~\cite{zhong2023ctg,strelgen2026}.

\begin{figure}[t]
\centering
\includegraphics[width=\columnwidth]{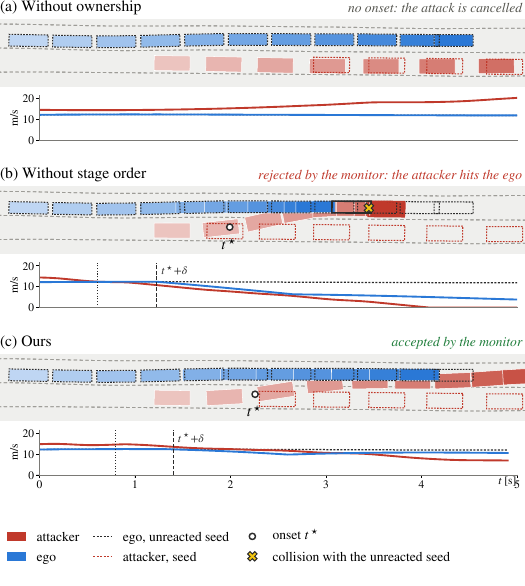}
\caption{A cut-in on US-101 generated from the same seed in three ways; under each panel the speeds of the generated
ego vehicle, the attacker and the unreacted seed (dashed), at the top right the monitor's verdict. (a) Without
ownership, the ego vehicle's clauses also move the attacker and cancel the attack. (b) Without the stage order, the
attacker is optimized against the responding ego vehicle: it chases the response, cuts in and brakes to a standstill, the
unreacted seed collides with it, and the monitor rejects the output, since the specification asks for an unsafe gap and
not a collision. (c) Ours: the attacker cuts in against the unreacted seed and slows down ahead of it, and the ego
vehicle keeps its seed until the release $t^\star+\delta$ and then brakes at \SI{2.8}{m/s^2}. Footprints every
\SI{0.5}{s}, light to dark; dashed outlines: the seeds.\label{fig:problem}}
\end{figure}

A further problem concerns validity (Fig.~\ref{fig:problem}). Guided diffusion models that generate
all agents, the ego vehicle included, optimize the terms of the attacker and of the ego vehicle in one
objective~\cite{zhong2023ctg,zhong2023ctgpp}. This objective does not represent the roles of the scenario. A
safety term of the ego vehicle then sends its gradient to the attacker as well and moves it back, which
cancels the attack. The attack is defined with respect to the ego vehicle that has not yet
reacted, but it is optimized against the ego vehicle that is responding. As a result, the attacker chases the response, and a cut-in can become a collision with the unreacted ego vehicle. Generators that run the ego vehicle as an
external planner in closed loop~\cite{chang2024safesim,lin2025ccdiff,ldscene2025} avoid these
interactions, but they produce no trajectory of the ego vehicle that could serve as a reference or a
label. Optimizing the attack first and a response afterwards has been used to find failures of a given
planner and their solutions~\cite{rempe2022strive,wang2023a2cost}. In guided generation, which terms may
move which agent and in which order have so far been designed for a given pair of roles or scenario
type~\cite{li2025omega,jing2026counterscene}.

In each of these problems, the objective does not represent which agent attacks, which agent responds,
and from which time on it may respond. We therefore state each
scenario as a role-typed, causal STL specification (Fig.~\ref{fig:overview}). Every clause names the
agent it may change and the agents it may read, and the clauses of the ego vehicle may read the attacker
only from the attack onset on. An LLM drafts the specification from a description, and a type checker
returns errors for repair. From the checked specification, a compiler derives how it is solved: each
clause moves only its owner, the attacker is solved against the unreacted seed of the ego vehicle and
fixed before the response is solved, and the monitor of the same specification decides acceptance. The
solver edits a seed scene (a diffusion sample or a planner's trajectory).

The contributions of this paper are the following.
\begin{enumerate}
\item We introduce a role-typed, causal fragment of signal temporal logic for adversarial scenarios, in
which every clause has an owner role and a causal type and clauses compose without weights.
Specifications are drafted by an LLM from a description and accepted by a type system that enforces
ownership and causality and drives a repair loop. A new scenario type needs a description and no new
weights. The rule that binds the attacker and the judge of the type remain fixed per type.
\item We derive a role-structured solve from the specification: the ownership of every clause, the order
of the roles, which solves the attack against the unreacted ego vehicle before the response, the attack
onset that releases the response, and the monitor as the acceptance gate. The response can then neither
cancel the attack nor draw it toward itself, and the monitor marks which outputs satisfy the specification.
\item We make the compiled specification efficient to solve at scale. The compiled objective is one tensor
program per specification, so the seeds of many windows are solved in one batch on an accelerator; a pre-check
skips seeds that cannot satisfy the specification, the monitor ends accepted roles early, and each distinct
accepted seed is a further scenario from the same input. Under a fixed budget this yields about nine times as many
solved windows per hour as one window at a time, and more valid scenarios per recorded input.
\end{enumerate}
We use the generated scenarios in both ways. Learned planners fail on them more often than on their
seeds. Fine-tuning a learned planner on them lowers its failure rate on critical scenarios more than
fine-tuning on the unedited seeds. One clause of the label sets the time headway of the response, and that headway
sets the size of the gain.

The remainder of this paper is organized as follows. Section~\ref{sec:related} reviews related work.
Section~\ref{sec:problem} formalizes the scenario and the generation problem, and
Section~\ref{sec:method} presents the method. Section~\ref{sec:experiments} reports the experiments,
Section~\ref{sec:discussion} discusses limitations, and Section~\ref{sec:conclusion} concludes the
paper.

\section{RELATED WORK}\label{sec:related}

\subsection{SAFETY-CRITICAL SCENARIO GENERATION}\label{sec:related:generation}

Surveys group the generation of safety-critical scenarios into data-driven, adversarial and
knowledge-based methods~\cite{ding2023survey} and review the use of foundation
models~\cite{foundationmodels2025survey}. Adversarial methods change recorded scenes against a planner,
by search over kinematic parameters~\cite{wang2021advsim}, by optimization in the latent space of a
learned traffic prior~\cite{rempe2022strive}, or by gradients through a differentiable bicycle
model~\cite{hanselmann2022king}. On CommonRoad, the other traffic participants of a scene are edited so
that the drivable area of the ego vehicle, computed by reachability analysis,
shrinks~\cite{althoff2018automatic,klischat2019evolutionary,klischat2020scenariofactory}, which makes
the criticality independent of a particular planner. Diffusion-based generators add an adversarial term to guided denoising~\cite{chang2024safesim,advdiffuser2024}
or encode a causal priority among agents~\cite{lin2025ccdiff}, and a flow-matching generator projects the sample onto collision constraints at each step~\cite{ccfm2026}. Whether a generated scene is fair to the planner under test is addressed by
optimizing a solution trajectory~\cite{rempe2022strive}, by feasibility constraints from reachable
sets or safety margins~\cite{safer2026,scenepilot2026}, and by attributing fault against a reference
driver~\cite{cars2026}. Most of these methods guide one designated attacker and run the ego vehicle as
an external planner in closed loop, and their objectives are written for each scenario category. We
use avoidability, judged independently, as a condition of validity in our evaluation. STRIVE optimizes
the attack against a planner and a solution for the ego vehicle in two separate
optimizations~\cite{rempe2022strive}. A2CoST searches for a collision and then for a collision-free
remedy of the ego vehicle with the adversaries fixed, from scenarios described in answer set
programming~\cite{wang2023a2cost}. AnchorDrive rolls out the ego vehicle and an attacker with an LLM
driver, checks the rollout with an LLM assessor, and then regenerates all agents with anchor-guided
diffusion~\cite{jiang2026anchordrive}. Guided diffusion models that generate the ego vehicle as well
optimize the terms of all agents together~\cite{zhong2023ctg,zhong2023ctgpp}. OMEGA lets the ego
vehicle and the attacker alternate best responses to each other inside guided diffusion until they
converge~\cite{li2025omega}. CounterScene applies the guidance gradients only to a causally
identified adversary while the planner under test reacts in closed loop~\cite{jing2026counterscene}. In
our method, the roles of the specification determine which terms may move which agent and the order of
the stages, for any scenario type and in one pass: the attacker against the unreacted seed of the ego
vehicle, then the response with the attacker fixed. Table~\ref{tab:positioning} summarizes these
differences.

\begin{table}[t]
\caption{Attack and Response in the Closest Methods\label{tab:positioning}}
\centering\footnotesize\setlength{\tabcolsep}{2pt}
\begin{tabular}{@{}>{\raggedright\arraybackslash}p{0.245\columnwidth}>{\raggedright\arraybackslash}p{0.17\columnwidth}>{\raggedright\arraybackslash}p{0.18\columnwidth}>{\raggedright\arraybackslash}p{0.16\columnwidth}>{\raggedright\arraybackslash}p{0.17\columnwidth}@{}}
\toprule
Method & Ego trajectory & Terms move & Attack against & Accepted by \\
\midrule
CTG, CTG++~\cite{zhong2023ctg,zhong2023ctgpp} & generated & all guided agents & reacting ego & none \\
STRIVE~\cite{rempe2022strive} & planner, then generated & all agents & planner rollout & solution found \\
A2CoST~\cite{wang2023a2cost} & planner, then generated & adversaries & planner & remedy found \\
Orchestration~\cite{rong2026orchestration} & planner & one actor & planner & constraints satisfied \\
AnchorDrive~\cite{jiang2026anchordrive} & LLM driver, then generated & all agents & reacting ego & LLM assessor \\
OMEGA~\cite{li2025omega} & generated & ego, attacker in turn & reacting ego & none \\
CounterScene~\cite{jing2026counterscene} & planner & adversary only & planner & none \\
Ours & generated & owner of each clause & unreacted seed & specification monitor \\
\bottomrule
\end{tabular}
\par\smallskip\parbox{\columnwidth}{\footnotesize Ego trajectory: whether the method outputs a generated ego trajectory. Terms move: which agents the objective's gradient or search changes. Attack against: the ego that the attack objective reads. Accepted by: the test that marks an output as a scenario, or none.}
\end{table}

\subsection{FORMAL SPECIFICATIONS AND LANGUAGE INTERFACES}\label{sec:related:specifications}

Scenarios are synthesized directly from formal specifications. In~\cite{klischat2020synthesizing}, a
specification is a sequence of scenes, each a set of predicates such as a vehicle being on a lanelet or
behind another vehicle. The trajectories of all vehicles follow from one mixed-integer quadratic
program, whose feasibility also decides whether the specification is satisfiable.
In~\cite{finkeldei2023synthesizing}, reachability analysis removes the binary variables and admits
nonlinear specifications such as traffic rules, and in~\cite{klischat2023falsifying}, sets of
specification-compliant trajectories are searched for failures of the vehicle under test. Scenario
Factory 2.0 combines these methods in one pipeline for CommonRoad~\cite{finkeldei2025scenariofactory2}.
Scenic describes scenes in a probabilistic programming language and falsifies a system against
them~\cite{fremont2019scenic}. These methods synthesize the traffic around a vehicle under test that is
executed separately, and they characterize criticality with planner-independent measures, for which
CommonRoad provides a toolbox~\cite{lin2023crime}. Traffic rules have been formalized in temporal logic
over similar predicates~\cite{maierhofer2020formalization}, and the robustness of STL specifications ordered by
priority is the objective of a minimum-violation motion planner for the ego vehicle~\cite{halder2026lexicographic}. Our method differs in its output, a joint
trajectory that includes a responding ego vehicle. We therefore compare against guided generators of
all agents and not against these tools.
For learned generators, STL serves as a guidance objective: CTG guides a diffusion model with
hand-written STL rules~\cite{zhong2023ctg}, STRELGen guides with spatio-temporal
logic whose robustness is pooled over all agents~\cite{strelgen2026}, and DIFF-MA guides a multi-robot
planner with one formula per agent~\cite{hetstl2026}. Language interfaces turn a request into
structured scene descriptions~\cite{tan2024lctgen}, Scenic programs~\cite{zhang2024chatscene},
constraints that an SMT solver satisfies around a planner under test, with triggers that start an
actor's maneuver~\cite{rong2026orchestration}, or loss code for guidance, which an LLM writes and, in
LD-Scene, debugs by running it~\cite{zhong2023ctgpp,ldscene2025}. Translating natural language into
temporal logic with an LLM, with a check or a repair loop, is studied in nl2spec~\cite{cosler2023nl2spec},
NL2TL~\cite{chen2023nl2tl} and, for traffic rules, TR2MTL~\cite{tr2mtl2024}. ARISE repairs LLM-written
Scenic scripts by compiling and simulating them~\cite{poddubnyy2026arise}. Our type checker adds the
ownership and causality rules of the fragment to such a loop. Of these, Orchestration is the nearest to
our specification: its constraints are satisfied exactly, without weights, and a trigger condition starts
the maneuver of the generated actor as our observation clause starts the response. They are solved for
one actor around a planner that is executed separately, the drafted constraints carry no owner or causal
type and pass no check before they are solved, and no learned traffic prior supplies the
scene~\cite{rong2026orchestration}. The works above either synthesize traffic around a separately executed
vehicle under test or compose objectives over all guided agents. We add to each clause the agent it may change
and the agents it may read, restrict the ego vehicle's access to the attacker in time, and apply the
specification as an edit of a seed scene that includes the ego vehicle.

\subsection{CAUSALITY AND REACTIVITY IN GENERATION}\label{sec:related:causality}

Closed-loop simulation keeps simulated agents non-anticipative by separating the vehicle under test
from the other agents and re-planning from the executed history~\cite{montali2023wosac}, and the
reactivity of such a rollout depends on its re-planning rate~\cite{liu2025rollingahead,scenediffuser2024}.
That conditioning on another agent's future lets a model react to it in advance has been measured for
conditional behavior prediction~\cite{tang2022interventional} and observed for a solution optimized over
the full horizon~\cite{rempe2022strive}. Fixing part of a sample by replacing it with known values is the
inpainting route to conditioning~\cite{lugmayr2022repaint,janner2022diffuser,scenediffuser2024}. We use it
to keep the generated ego vehicle at its seed until an onset that the specification defines.

\section{PROBLEM FORMULATION}\label{sec:problem}

This section defines the scene and the seed that the method edits, the role-typed specification of a
scenario, causality and avoidability, and the generation problem.

\subsection{SCENE, SEED, AND ROLES}\label{sec:problem:scene}

Let us denote by $\Agents=\{1,\dots,A\}$ the agents of a scene. Time is discrete with step length
$\Delta t$, the states at $t=0$ are given, and the method generates the steps $t=1,\dots,T$. The horizon is thus $T$ steps or $T\Delta t$ seconds. Agent $i$
has the state $s_i(t)=(x_i(t),y_i(t),v_i(t),\theta_i(t))$ with position, speed and heading, and the
action $u_i(t)=(a_i(t),\omega_i(t))$ with longitudinal acceleration and yaw rate. Bold symbols denote
sequences, $\bu_i=(u_i(0),\dots,u_i(T-1))$ and $\bx_i=(s_i(1),\dots,s_i(T))$, and
$\bx=(\bx_i)_{i\in\Agents}$ is the joint trajectory. We model each agent that the method changes as a
unicycle, integrated with the midpoint rule,
\begin{equation}\label{eq:unicycle}
\begin{aligned}
v_i(t+1)&=v_i(t)+a_i(t)\,\Delta t,\\
\theta_i(t+1)&=\theta_i(t)+\omega_i(t)\,\Delta t,\\
x_i(t+1)&=x_i(t)+\bar v_i\cos\bar\theta_i\,\Delta t,\\
y_i(t+1)&=y_i(t)+\bar v_i\sin\bar\theta_i\,\Delta t,
\end{aligned}
\end{equation}
where $\bar v_i$ and $\bar\theta_i$ are the means of speed and heading at $t$ and $t+1$. The actions
lie in a set $\Uset$ of action sequences whose rollout respects bounds on the acceleration, the yaw rate,
the lateral acceleration, the speed and the curvature. We write the rollout of~\eqref{eq:unicycle} as
$\bx_i=F(s_i(0),\bu_i)$, which is differentiable in $\bu_i$.

The method does not generate a scene from scratch. It edits a given joint trajectory.
\begin{definition}[Seed]\label{def:seed}
A seed is a joint trajectory $\xseed$ of one scene over the horizon, together with a designated ego
vehicle. For each agent $i$ that the method may change, the seed actions $\useed_i\in\Uset$ are
obtained by fitting $F(s_i(0),\cdot)$ to $\xseed_i$, and $\xseed_i$ is replaced by
$F(s_i(0),\useed_i)$.
\end{definition}
A seed can be a sample of a generative model or a recorded scene in which a planner provides the ego
vehicle's trajectory. Section~\ref{sec:exp:setup} gives the sources we use.

A scenario assigns a role to each agent through a map $r:\Agents\to\Roles$ with
$\Roles=\{\att,\ego,\by,\oth\}$. The attacker ($\att$) creates the threat, and the ego vehicle
($\ego$) is the vehicle under test. A bystander ($\by$) is any other agent that the method may change,
and a context agent, of the role $\oth$, keeps its seed and is only read. We write $\Agents_q$ for the
agents of role $q$. The roles that the method changes are ordered by the priority $\att\succ\ego\succ\by$,
and the generation problem of Section~\ref{sec:problem:generation} is instantiated in this order: the attacker is solved first with the ego vehicle at its seed, the ego vehicle then responds
to the solved attacker, and bystanders adapt to both. We call the ego vehicle at its seed the
\emph{unreacted ego vehicle}. In this paper, each scenario has one attacker and one ego vehicle.
\label{sec:problem:roles}

\subsection{ROLE-TYPED SPECIFICATION}\label{sec:problem:specification}

A scenario is specified in STL~\cite{maler2004stl} with role annotations. Each predicate
$\mu(\bx,t)$ is a real-valued function of the states of the agents bound to its role arguments, in a
physical unit. For example, $\pred{gap\_lon}(\ego,\att;L)$ is the arc length along lane $L$ from the
front of the ego vehicle to the rear of the attacker. Formulas follow the grammar
\begin{equation}\label{eq:grammar}
\begin{aligned}
\varphi ::={}& \mu>c \mid \mu<c \mid \neg\varphi \mid \varphi_1\wedge\varphi_2 \mid
\varphi_1\vee\varphi_2 \mid \Gop_{I}\varphi \\
& \mid \Fop_{I}\varphi \mid \varphi_1\,\Uop_{I}\,\varphi_2 \mid \Hop_{I}\varphi \mid \Oop_{I}\varphi ,
\end{aligned}
\end{equation}
where $\mu$ may also stand for $|\mu|$, $\mu<c$ abbreviates $\neg(\mu>c)$, and the bounded future
operators always ($\Gop$), eventually ($\Fop$) and until ($\Uop$) and the past operators historically
($\Hop$) and once ($\Oop$) range over an interval $I=[\alpha,\beta]$. A bound is a time in seconds, which is
rounded to steps, or is anchored at the attack onset $t^\star$ defined below and then written
$t^\star+\kappa$ with $\kappa\ge0$; in an interval, $T$ denotes the end of the horizon. All times in the
specification tables are given in seconds and rounded to steps.

Formulas are evaluated on $\bx$ and indexed by the generated steps: $\rho(\varphi,\bx,\tau)$ for
$\tau=0,\dots,T-1$ refers to the state $s(\tau+1)$. The time origin of a formula is therefore the first
generated step. We use the robustness of~\cite{donze2010robustness,fainekos2009robustness}, with each
atom normalized by a nominal scale of its predicate. The normalization preserves the sign: an atom $\mu>c$ has
$\rho=(\mu(\bx,\tau)-c)/\bar\mu$ with $\bar\mu>0$, negation flips the sign, $\wedge$ and $\Gop$ take
the minimum, $\vee$ and $\Fop$ the maximum, $\Hop$ and $\Oop$ likewise over the past interval
$[\tau-\beta,\tau-\alpha]$, and
\begin{equation}\label{eq:until}
\begin{aligned}
&\rho(\varphi_1\Uop_{[\alpha,\beta]}\varphi_2,\bx,\tau)\\
&\quad=\max_{\tau'\in[\tau+\alpha,\tau+\beta]}
\min\Bigl(\rho(\varphi_2,\bx,\tau'),\min_{\tau\le\tau''<\tau'}\rho(\varphi_1,\bx,\tau'')\Bigr),
\end{aligned}
\end{equation}
where $\varphi_1$ is required on the half-open prefix $[\tau,\tau')$, so that $\varphi_1$ and
$\varphi_2$ may exclude each other, as in the clause $\cl{avoidable}$ of the Appendix, and an empty
prefix imposes nothing. An interval is intersected with the horizon. Over an empty interval, $\Gop$ and
$\Hop$ are vacuously true and $\Fop$ and $\Oop$ false, with robustness $\pm\infty$, a large constant in
the implementation. A predicate over a role that holds several agents, such as the context agents, is
read for each agent of the role that is present at the step. Its robustness is the minimum over
them, a universal quantification whose negation is the existential one. We take $\rho(\varphi,\bx,0)\ge0$
as satisfaction of $\varphi$, which differs from the Boolean semantics only on the boundary
$\rho=0$~\cite{fainekos2009robustness}. Since the contact clauses carry a margin, the boundary is never a
contact.\label{def:robustness}

\begin{definition}[Role-typed specification]\label{def:specification}
A role-typed specification is a set of clauses
$\Phi=\{(\varphi_k,\own_k,\reads_k,\kind_k)\}_{k=1}^{n_\Phi}$. The owner $\own_k$ is the one role whose
trajectories the clause may change, and the read set $\reads_k$ lists the roles it may observe without
changing them. Every role that $\varphi_k$ references belongs to $\{\own_k\}\cup\reads_k$. The kind
$\kind_k\in\{\mathrm{attack},\mathrm{require},\mathrm{observe},\mathrm{check}\}$ marks the clause that
states the attack, any other requirement, the observation clause $\psiobs$, and a check, which is verified
but never optimized. In addition, the specification states the priority of its roles, a reaction delay $\delta$
of the ego vehicle and, for each role $q$, a hold time $h_q\ge0$ before which the role keeps its seed
actions, with $h_q=0$ unless stated.
\end{definition}
We write $\Phi_q\subseteq\{1,\dots,n_\Phi\}$ for the indices of the clauses owned by role $q$, other than
the observation clause. The observation clause defines when the attack becomes observable to the ego
vehicle.

\begin{definition}[Attack onset and release]\label{def:onset}
The observation clause has the formula $\psiobs$, is owned by the attacker, references no other role, and
carries a sustain time $\sigma$ in steps. The attack onset $t^\star$ of a trajectory $\bx$ is the earliest
step $t$ with $\rho(\psiobs,\bx,t')\ge0$ for $t'=t,\dots,t+\sigma-1$, $t+\sigma\le T$ and
$t+\delta\le T-1$, where $\delta$ is counted in steps. The release is $\trel=t^\star+\delta$. If no such
$t$ exists, the trajectory has no onset.
\end{definition}
The sustain condition is $\Gop_{[0,\sigma-1]}\psiobs$ at the step level. Since $\psiobs$ references only
the attacker, $t^\star$ is the same on $\bx$ and on any view of $\bx$ in which the other agents differ
(Section~\ref{sec:problem:generation}). The condition $t+\delta\le T-1$ excludes a release whose
intervals are empty.

\subsection{CAUSALITY AND AVOIDABILITY}\label{sec:problem:causal}

A clause of the ego vehicle may depend on the attacker only once the attack is observable, and the attack
must be avoidable by a driver who reacts from then on. We state the first for the specification and the
second for the generated scenario.
\begin{definition}[Causal clause]\label{def:causal}
Let a formula be evaluated at a step $t$ of which it is known that $t\ge t^\star+\ell$, with
$\ell\in\mathbb{Z}\cup\{-\infty\}$ and $\ell=-\infty$ when nothing is known. The steps that its
subformulas may read are bounded below by the recursion: a predicate is read at $t$ itself, so it may be
read before the onset if and only if $\ell<0$; $\neg$, $\wedge$ and $\vee$ pass $\ell$ to their arguments;
$\Gop_{[\alpha,\beta]}$ and $\Fop_{[\alpha,\beta]}$ pass $\ell+\alpha$ if $\alpha$ is a constant and
$\kappa+\max(0,\ell)$ if $\alpha=t^\star+\kappa$; $\varphi_1\Uop_{[\alpha,\beta]}\varphi_2$ passes
$\ell$ to $\varphi_1$ and the bound of its interval to $\varphi_2$; and $\Hop_{[\alpha,\beta]}$ and
$\Oop_{[\alpha,\beta]}$ pass $\ell-\beta$, and $-\infty$ if $\beta$ is anchored at the onset. A clause
owned by the ego vehicle is causal if, starting from $\ell=-\infty$ at the clause itself, no predicate in
it that references another role may be read before the onset. A specification is causal if every clause
owned by the ego vehicle is causal.
\end{definition}
Because the condition is syntactic, the type checker decides it (Section~\ref{sec:method:typecheck}). For
example, $\Gop_{[t^\star+\delta,T]}\,\pred{box\_sep}(\ego,\att)>\SI{0.05}{m}$ is causal, and
$\Gop_{[0,T]}\,\pred{box\_sep}(\ego,\att)>\SI{0.05}{m}$ is not, since it couples the ego vehicle to
the attacker before the attack is observable. A past interval under an onset-anchored bound,
$\Gop_{[t^\star+\kappa,T]}\Oop_{[0,w]}\,\mu$, is causal if and only if $\kappa\ge w$. With $\kappa<\delta$, a causal clause
constrains steps at which the ego vehicle is still held at its seed
(Section~\ref{sec:problem:generation}); the clauses of the four specifications in the Appendix use $\kappa=\delta$.

\begin{definition}[Avoidability]\label{def:nonanticipation}\label{def:avoidable}
A generated scenario with release $\trel$ is avoidable if an ego trajectory
$\bx_{\ego}=F(s_{\ego}(0),\bu_{\ego})$ with $\bu_{\ego}\in\Uset$ exists that keeps the seed actions,
$u_{\ego}(t)=\useed_{\ego}(t)$ for $t=0,\dots,\trel-1$, is in contact with no other agent, except when struck from
behind ($\pred{box\_sep}^{\mathrm{rear}}$ of the Appendix), and stays on the road.
\end{definition}
The seed is drawn before the attack is edited into the scene and carries no information about the
attack. Such a trajectory therefore shows that a driver with the reaction delay $\delta$ can avoid it. This definition asks only for the existence of the trajectory. The witness that the method produces
(Section~\ref{sec:problem:generation}) is computed with the attacker's trajectory after the release known.
For this reason, Section~\ref{sec:exp:setup} also checks avoidability with a fixed braking response that uses no
such knowledge.

\subsection{GENERATION PROBLEM}\label{sec:problem:generation}

Generation is a minimal modification of the seed, solved for one role after the other in the priority
order, as the criticality-increasing methods of the CommonRoad ecosystem edit recorded scenes to make
them critical~\cite{althoff2018automatic,klischat2019evolutionary}, now integrated in Scenario
Factory~2.0~\cite{finkeldei2025scenariofactory2}. The modification of agent $i$ is measured by
\begin{equation}\label{eq:proximity}
d(\bx_i,\xseed_i)=\frac{1}{T}\sum_{t=1}^{T}\Bigl(\|\Delta p_i(t)\|^2+\Delta v_i(t)^2\Bigr),
\end{equation}
where $\Delta p_i$ and $\Delta v_i$ are the deviations of the position $p_i=(x_i,y_i)$ and of the speed
from the seed, with unit weights, so that a speed deviation of \SI{1}{m/s} counts as much as a position
deviation of \SI{1}{m}. A clause of role $q$ is evaluated on the staged view $\bx^{(q)}(\bu)$, in
which the roles of higher priority are at their solutions, the agents of $q$ at the trajectories
$F(s_i(0),\bu_i)$ under optimization, and all other agents at their seeds.

\begin{definition}[Admissible seed]\label{def:admissible}
A seed is admissible if its ego trajectory is in contact with no agent, except when struck from behind, and stays on
the road over the horizon, that is, if the seed with $\trel=T$ satisfies Definition~\ref{def:avoidable}.
\end{definition}

\begin{problem}[Minimal modification for role $q$]\label{prob:minmod}
Given an admissible seed and the solutions of the roles of higher priority, find
\begin{equation}\label{eq:pq}
\begin{aligned}
\min_{\bu_i\in\Uset,\ i\in\Agents_q}\quad & \sum_{i\in\Agents_q} d\bigl(F(s_i(0),\bu_i),\xseed_i\bigr)\\
\text{s.t.}\quad & \rho\bigl(\varphi_k,\bx^{(q)}(\bu),0\bigr)\ge0 \quad \forall k\in\Phi_q,\\
& u_i(t)=\useed_i(t) \quad t<h_q,\ i\in\Agents_q,\\
& u_{\ego}(t)=\useed_{\ego}(t) \quad t<\trel \ \text{ if } q=\ego,
\end{aligned}
\end{equation}
where $t^\star$, and with it $\trel$, is the onset of the attacker's solution. If the attacker's
solution has no onset, the problem for the ego vehicle is not posed, and the output is rejected.
\end{problem}
We refer to~\eqref{eq:pq} as (P$_q$). The clauses of kind check are constraints that the solver verifies
and never optimizes. Since the attacker's clauses are evaluated in $\bx^{(\att)}$, they read the
unreacted ego vehicle. The attack is judged against the unreacted ego vehicle it is aimed at, and the
response, which may defuse the attack, is judged by the ego vehicle's own clauses. For lead-vehicle
braking, the attack clause requires that the unreacted ego vehicle would hit the braking lead vehicle.
The last constraint of (P$_{\ego}$) is the prefix condition of Definition~\ref{def:avoidable}. If the
seed is admissible, the attacker's clauses exclude contact with the unreacted ego vehicle before $\trel$
(the clause $\cl{no\_hit}$ of the cut-in specification over the horizon, and $\cl{no\_early\_hit}$ of the VRU and LTAP specifications up to \SI{1.8}{s}, in the Appendix),
and the ego vehicle's clauses entail no contact, except from behind, and a footprint on the road from
$\trel$ on, then an accepted ego trajectory is a witness of
avoidability. Admissibility is required because the ego vehicle's clauses start at $\trel$ and never
examine the held prefix.

A generated trajectory is accepted if it has an onset and every clause $k\in\Phi_q$ of every role $q$
holds on the staged view, $\rho(\varphi_k,\bx^{(q)},0)\ge0$. The Boolean evaluation of these conditions is
the monitor of $\Phi$.\label{def:acceptance}
Acceptance is the method's own verdict. Whether an accepted scenario is valid is decided in
Section~\ref{sec:experiments} by judges that share no code with the specification.

\section{METHOD}\label{sec:method}

\begin{figure*}[t]
\centering
\includegraphics[width=\textwidth]{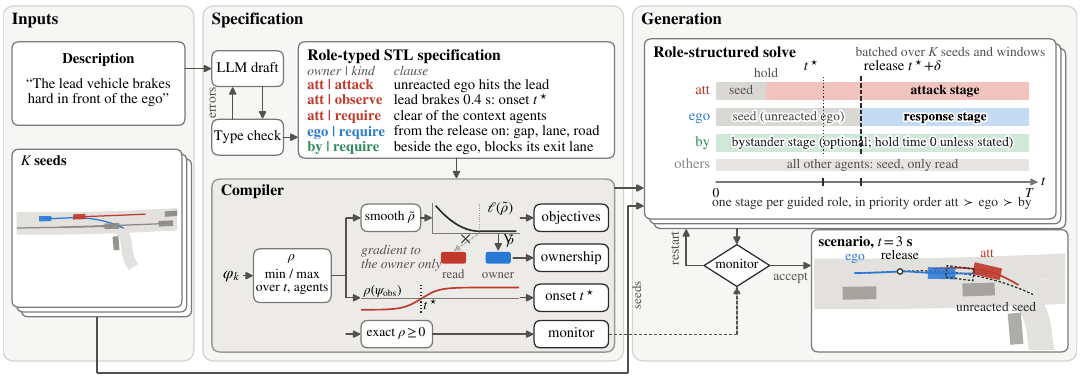}
\caption{Overview of the method for the description of a lead-vehicle braking scenario. Middle: an LLM drafts a role-typed
specification, a type check returns errors for repair (contribution 1), and the compiler turns every clause $\varphi_k$
into four objects. The exact robustness $\rho$, a minimum and maximum over time and agents, yields
the monitor's verdict $\rho\ge0$ and, for the observation clause, the onset $t^\star$. Its smooth lower
bound $\tilde\rho$ enters a hinge loss whose gradient moves only the clause's owner. Right: the solver runs in stages
derived from the roles (contribution 2): the attacker keeps its seed during its hold time and is then edited against
the unreacted seed of the ego, then the ego's
response is edited with the attack fixed, and the ego keeps its seed until the release $t^\star+\delta$. A
specification may name a further guided role, such as a bystander that blocks the ego's escape lane: it owns
its clauses, whose gradients move only that role, and is solved in its own stage after the ego's. All other agents
are context agents and keep their seed trajectories. The
monitor of the compiled specification accepts the result or restarts the role from maneuver primitives, and the $K$ seeds of many windows are solved in one batch
(contribution 3). Thumbnails (nuScenes, bird's-eye view, true vehicle sizes): left, a seed scene with every agent's
trajectory (boxes at $t=0$); bottom right, the generated scenario at $t=3$\,s: the lead vehicle brakes into the
path of the ego's unreacted seed (dashed), and the generated ego, identical to its seed until the release
(circle), stays behind the lead vehicle.\label{fig:overview}}
\end{figure*}

Fig.~\ref{fig:overview} shows the method. A specification is drafted by an LLM and type-checked
(Sections~\ref{sec:method:language} and~\ref{sec:method:typecheck}). From the specification, the compiler derives a
role-structured solve: the objectives, the ownership of each clause, the attack onset and the monitor
(Section~\ref{sec:method:compile}). It also derives the order of the roles, in which the attacker is solved against
the unreacted seed of the ego vehicle and fixed before the response is solved
(Section~\ref{sec:method:staged}).
Section~\ref{sec:method:efficiency} describes the mechanisms that reduce the computation.

\subsection{SPECIFICATION LANGUAGE}\label{sec:method:language}

We state scenarios in STL for three reasons. First, STL is defined over real-valued signals~\cite{maler2004stl},
which~\cite{stlcgpp2025} names as its advantage over logics of discrete propositions. Its atoms are
therefore predicates of the trajectory in physical units, such as a gap in meters. Through its quantitative semantics, STL measures how well a trajectory satisfies a requirement~\cite{donze2010robustness,stlcgpp2025}.
Since this robustness is differentiable almost everywhere, it serves as a guidance objective for gradient-based optimization and
for diffusion models~\cite{leung2020stlcg,zhong2023ctg,hetstl2026}. Its sign is the Boolean verdict of
the same formula, which makes the objective and the acceptance test two readings of one specification.
Robustness is also defined for metric temporal logic over metric spaces~\cite{fainekos2009robustness}, and STL
states the predicates directly on the signals that the solver changes. Second, the temporal operators of
STL carry metric time intervals, which linear temporal logic lacks. With them, the time at which an attack
becomes observable and the time from which the ego vehicle may respond are part of the specification,
and the
causality of a clause becomes a property of its syntax (Section~\ref{sec:method:typecheck}). Anchoring an
interval at the onset extends the constant bounds of STL. It binds a time that the formula itself defines,
in the spirit of the freeze quantifier of timed propositional temporal logic~\cite{alur1994tptl}. Third,
conjunction and disjunction are minima and maxima, and clauses therefore compose without weights.

Specifications are written in the STL of~\eqref{eq:grammar} over one library of driving predicates
(Table~\ref{tab:predicates}), the same for all scenario families and data sets. Each predicate is a
differentiable function of the states of its agents, in a physical unit, with a nominal scale $\bar\mu$.
Lanes are referenced relative to roles: $\mathrm{lane}(\ego)$ is the chain of lanes that the ego
vehicle's seed follows through forks and junctions, $\mathrm{left\_of}(\cdot)$ and
$\mathrm{right\_of}(\cdot)$ are adjacent lanes, a turn path such as
$\mathrm{left\_turn\_of}(\mathrm{lane}(\att))$ is resolved from the junction connectivity of the map,
and $\mathrm{road}$ is the road corridor with its local width. A specification therefore names no lane of
a particular map, and since time is stated in seconds, one specification runs on any scene and step
length. Contact is tested with $\pred{box\_sep}$, because the lane-frame predicates distort away from
the center line of a curved lane.

\begin{table}[t]
\caption{Predicate Library\label{tab:predicates}}
\centering
\footnotesize
\setlength{\tabcolsep}{2pt}
\begin{tabular}{@{}>{\raggedright\arraybackslash}p{0.38\columnwidth}>{\raggedright\arraybackslash}p{0.42\columnwidth}l@{}}
\toprule
Predicate & Value & Unit \\
\midrule
$\pred{gap\_lon}(i,j;L)$ & arc length along $L$ from $i$'s front to $j$'s rear & m \\
$\pred{lat\_off}(i;L)$ & lateral offset of $i$ from the center line of $L$ & m \\
$\pred{lat\_sep}(i,j;L)$ & lateral separation of the footprints across $L$ & m \\
$\pred{lat\_shift}(i;L)$ & lateral displacement of $i$ from its initial offset & m \\
$\pred{speed}(i)$, $\pred{acc}(i)$ & speed, longitudinal acceleration & m/s, \\ & & m/s$^2$ \\
$\pred{rss\_margin}(i,j;L,a_i)$ & $\pred{gap\_lon}$ minus a simplified RSS distance~\cite{shalevshwartz2017rss} of follower $i$ braking at $a_i$, with a floor of \SI{2}{m} (Appendix) & m \\
$\pred{box\_sep}(i,j)$ & signed separating-axis distance of the oriented footprints; $\pred{box\_sep}^{\mathrm{rear}}$ excuses $j$ when it runs into $i$ from behind (Appendix) & m \\
$\pred{road\_margin}(i;L)$ & margin of $i$'s footprint inside the road around $L$ & m \\
\bottomrule
\end{tabular}
\end{table}

Table~\ref{tab:eb} shows the specification of lead-vehicle braking, our running example. The attack
clause $\cl{brake}$ requires the lead vehicle to brake harder than \SI{3.5}{m/s^2} for \SI{1}{s},
starting between \SI{0.8}{s} and \SI{2}{s}, so that the unreacted ego vehicle would hit it. Until
\SI{0.8}{s}, the lead vehicle keeps its seed, and $\cl{quiet}$ checks that it does not already brake
there, so that the scene has a part before the attack. A clause of kind check is evaluated by the
monitor and never optimized, because it states a property of the part of the trajectory that the solver
holds. The observation clause holds when the lead
vehicle's deceleration exceeds \SI{2}{m/s^2} for \SI{0.4}{s}, four steps of \SI{0.1}{s}, which is the
interval $\Gop_{[0,0.3]}$ of $\cl{quiet}$. From the
release on, the ego vehicle's clauses require it to stay behind the lead vehicle, in its lane and on the road, without contact
with third parties.
The constants of the four specifications in the library were set during development on windows of the full case sets of
both platforms, before the final version of the method was fixed and before the cases of Section~\ref{sec:experiments} were
drawn. Since they describe the scenario type, they hold on every platform and are the same in all methods
that use the specification. Hard braking is a deceleration beyond \SI{3.5}{m/s^2}.
Observable braking is a deceleration beyond \SI{2}{m/s^2} sustained for \SI{0.4}{s}, because an
instantaneous threshold was triggered by the single-step jitter of the diffusion samples. The hold time of
\SI{0.8}{s} is longer than the reaction delay $\delta=\SI{0.6}{s}$, so that a part of the scene precedes
the attack. For drafted specifications, rule T5 of the type checker requires thresholds on speed and
acceleration to lie inside the action bounds. Apart from the severity constant (Section~\ref{sec:exp:severity}) and the
headway clause of the training labels (Section~\ref{sec:exp:rq6}), we did not vary these constants
(Section~\ref{sec:discussion}).

\begin{table*}[t]
\caption{Lead-Vehicle Braking Specification (Simplified)\label{tab:eb}}
\centering
\footnotesize
\setlength{\tabcolsep}{4pt}
\relpenalty=10000 \binoppenalty=0 %
\begin{tabular}{@{}llll>{\raggedright\arraybackslash}p{0.21\textwidth}>{\raggedright\arraybackslash}p{0.365\textwidth}@{}}
\toprule
Clause & Owner & Reads & Kind & Meaning & Formula \\
\midrule
\cl{brake} & $\att$ & $\ego$ & attack & hard braking into the unreacted ego vehicle &
$\Fop_{[0.8,2]}\Gop_{[0,1]}\,\pred{acc}(\att)<\SI{-3.5}{m/s^2}\wedge\Fop_{[2.4,T]}\,\pred{box\_sep}(\ego,\att)<0$ \\
\cl{quiet} & $\att$ & $\emptyset$ & check & no braking before the hold time; verified, not optimized & $\Gop_{[0,0.6]}\,\neg\Gop_{[0,0.3]}\,\pred{acc}(\att)<\SI{-2}{m/s^2}$ \\
\cl{att\_lane} & $\att$ & $\ego$ & require & attacker stays in the ego's lane & $\Gop_{[0.8,T]}\,|\pred{lat\_off}(\att;\mathrm{lane}(\ego))|<\SI{0.8}{m}$ \\
\cl{att\_road} & $\att$ & $\emptyset$ & require & attacker stays on the road & $\Gop_{[0.8,T]}\,\pred{road\_margin}(\att;\mathrm{road})>0$ \\
\cl{no\_third\_party\_hit} & $\att$ & $\oth$ & require & attacker clear of the context agents & $\Gop_{[0,T]}\,\pred{box\_sep}(\att,\oth)>\SI{0.3}{m}$ \\
\cl{observe} & $\att$ & $\emptyset$ & observe & onset $t^\star$: braking the ego vehicle can observe & $\pred{acc}(\att)<\SI{-2}{m/s^2}$, sustained for $\sigma=\SI{0.4}{s}$ \\
\cl{ego\_clear} & $\ego$ & $\att$ & require & ego vehicle stays behind the attacker, no contact &
$\Gop_{[t^\star+\delta,T]}\bigl(\pred{gap\_lon}(\ego,\att;\mathrm{lane}(\ego))>\SI{0.5}{m}\wedge\pred{box\_sep}(\ego,\att)>\SI{0.05}{m}\bigr)$ \\
\cl{ego\_lane} & $\ego$ & $\emptyset$ & require & ego vehicle stays in its lane & $\Gop_{[t^\star+\delta,T]}\,|\pred{lat\_off}(\ego;\mathrm{lane}(\ego))|<\SI{0.8}{m}$ \\
\cl{ego\_no\_third\_party\_hit} & $\ego$ & $\oth$ & require & ego vehicle clear of the context agents, unless struck from behind & $\Gop_{[t^\star+\delta,T]}\,\pred{box\_sep}^{\mathrm{rear}}(\ego,\oth)>\SI{0.3}{m}$ \\
\cl{ego\_road} & $\ego$ & $\emptyset$ & require & ego vehicle stays on the road & $\Gop_{[t^\star+\delta,T]}\,\pred{road\_margin}(\ego;\mathrm{road})>0$ \\
\bottomrule
\multicolumn{6}{@{}p{0.97\textwidth}@{}}{\rule{0pt}{2.5ex}Priority $(\att,\ego)$, context role $\oth$,
reaction delay $\delta=\SI{0.6}{s}$, horizon $T=\SI{5}{s}$, hold time $h_{\att}=\SI{0.8}{s}$, before which
the attacker keeps its seed, which $\cl{quiet}$ checks without optimizing. Simplified: the contact test of $\cl{brake}$
omits its lane-frame overlap conjunct; $\pred{box\_sep}^{\mathrm{rear}}$ excuses a vehicle that runs
into the braking ego from behind.}
\end{tabular}
\end{table*}

\subsection{DRAFTING AND TYPE CHECKING}\label{sec:method:typecheck}\label{sec:method:llm}

An LLM drafts the specification from a description in natural language. The prompt states the grammar of a
specification as a JSON object, the predicate table, the rules below in plain language, and a few
examples. Each agent of the scene is described by its lane offset, gap, speed and presence relative to the
ego vehicle. Since the predicate table in the prompt is generated from the compiler's library, the two
cannot disagree. A type checker then applies five rules:
\begin{enumerate}
\item[T1)] well-formedness: known predicates and references, intervals inside the horizon;
\item[T2)] ownership: the owner is a role the method changes, and every other role the formula
references is in the read set;
\item[T3)] causality: every clause of the ego vehicle is causal (Definition~\ref{def:causal}), decided
syntactically from the earliest step each operator can read, and there is exactly one observation
clause, owned by the attacker;
\item[T4)] non-vacuity: there is an attack clause, and no attack clause already holds on the seed;
\item[T5)] physical consistency: thresholds on speed and acceleration lie inside $\Uset$.
\end{enumerate}
Each violation returns a structured error that names the clause, the rule and the offending part, for
example that an ego clause reading the attacker over $[0,T]$ couples the ego vehicle to the attacker
before the onset. The errors are returned to the LLM, which answers with a corrected specification. After
at most three repair rounds, a specification that still fails is never executed. Iterating an LLM's output on feedback is used in reward-code synthesis~\cite{ma2024eureka}, and checking a specification before
synthesis is established for mixed-integer scenario synthesis~\cite{klischat2020synthesizing}. The
rules specific to the role-typed fragment are T2, T3 and the comparison with the seed in T4. Fixed attacker selection rules of each family bind the attacker. One drafted specification is thus bound in
turn to each candidate of a seed window.

\subsection{COMPILATION: OBJECTIVES, OWNERSHIP, ONSET, AND MONITOR}\label{sec:method:compile}

The compiler turns a checked specification and a role binding into four objects, which take the place
of a hand-written objective and its separate check.

\subsubsection{Objectives}
Each clause is put in negation normal form, and the smooth robustness $\rt$ replaces
the minimum and maximum of Section~\ref{sec:problem} by
\begin{equation}\label{eq:smooth}
\begin{aligned}
\widetilde{\min}_j v_j&=-\zeta\log\sum_j e^{-v_j/\zeta},\\
\widetilde{\max}_j v_j&=\sum_j \frac{e^{v_j/\zeta}}{\sum_l e^{v_l/\zeta}}\,v_j ,
\end{aligned}
\end{equation}
with $\zeta=0.1$ in normalized units. The smooth maximum is a softmax-weighted mean.
\begin{remark}[Lower bound]\label{prop:lowerbound}
For every formula in negation normal form, every $\bx$ and every $t$,
$\rt(\varphi,\bx,t)\le\rho(\varphi,\bx,t)$, since both operators in~\eqref{eq:smooth} are lower bounds
of the exact ones and the exact minimum and maximum are monotone in each argument. Hence
$\rt(\varphi,\bx,0)\ge0$ implies that the monitor accepts $\varphi$ on $\bx$.
\end{remark}
The usual log-sum-exp maximum is an upper bound, with which the optimizer could reach a smooth
robustness that the monitor rejects. Each clause enters the objective through the hinge
$\ell(\rt)=\max(0,h-\rt)^2$ with $h=0.1$, whose gradient vanishes once the clause holds with margin $h$.
Clauses are composed by minimum and maximum inside a clause and summed over clauses with multipliers
that start at one. The method thus has parameters, namely $\zeta$, $h$, the proximity weight and the step
sizes below, but they are the same for all scenario types and data sets, and no weight is set per type.
Differentiable robustness is available in existing libraries~\cite{stlcgpp2025}.

\subsubsection{Ownership}
Let $\Phi_q^{\mathrm{opt}}\subseteq\Phi_q$ be the clauses of kind attack or require; clauses of kind check are verified
by the monitor and not optimized. Each clause $k\in\Phi_q^{\mathrm{opt}}$ is instantiated once for each agent $i\in\Agents_q$ and
evaluated on
\begin{equation}\label{eq:detach}
\bx^{(i)}=\bigl(F(s_i(0),\bu_i),\ \sg(\bx^{(q)}_{-i})\bigr),
\end{equation}
where $\sg$ is the stop-gradient and $\bx^{(q)}_{-i}$ holds all other agents of the staged view.
\begin{remark}[Ownership]\label{prop:ownership}
For an agent $i$ of role $q$, the objective $J_i$ built from these instances satisfies
$\nabla_{\bu_j}J_i=0$ for every agent $j\neq i$, because $\bu_j$ enters $J_i$ only through
$\sg(\cdot)$ in~\eqref{eq:detach}.
\end{remark}
The clause $\cl{ego\_clear}$ reads the attacker, but its gradient reaches only the ego vehicle. The clause can therefore
slow the ego vehicle down and cannot move the lead vehicle away. A joint objective that keeps the
attacker differentiable sends the same gradient to the attacker and weakens the attack. Updating one
agent with the others fixed is the best-response structure of~\cite{stlgame2025}, and per-agent
specifications guide the planner of~\cite{hetstl2026}. The role-typed specification adds the assignment
of each clause instance to the one agent it updates, also for clauses that read several roles. It also
shapes how clauses are written. A disjunction whose larger disjunct depends only on a role the owner
cannot move gives the owner almost no gradient. For this reason, the library writes response clauses over
quantities
their owner controls, such as staying behind the lead vehicle along its own lane.

\subsubsection{Onset and Monitor}
The onset function evaluates $\psiobs$ exactly on the attacker's staged view and
returns $t^\star$ by Definition~\ref{def:onset}. The monitor evaluates each clause with the exact robustness on the staged view of its owner (Section~\ref{sec:problem}). Boolean checks without a
gradient, such as an at-fault rule, can be added to its verdict. Finally, the compiler evaluates the
attack clauses on the seed for the second part of T4.

\subsection{STAGE ORDER}\label{sec:method:staged}\label{sec:method:pinning}

The solver solves (P$_q$) for one role after the other in the priority order that the compiler derives
from the ownership: the attacker first, while all other agents keep their seeds, then the ego vehicle with
the attacker fixed, then the bystanders. A specification may give further roles clauses of their own, a bystander that blocks the ego vehicle's escape
lane for instance, and each is solved in its own stage after the ego vehicle's. With one attacker and one ego vehicle per scenario in this paper, all other agents are context agents:
the clauses read their trajectories, and the solver keeps them throughout. Because the attacker's stage
sees only the unreacted seed of the ego vehicle, the attack is not shaped by a response. The response,
in turn, cannot move the attacker back, since the ego vehicle's clauses move only the ego vehicle. Ownership alone does not provide the first
property. The attack clauses describe a counterfactual, namely what the unreacted ego vehicle would
experience, whereas the response clauses describe the ego vehicle that reacts. If both roles were solved
together, the attack clauses would read the reacting ego vehicle, and the attacker would chase the
response instead of attacking the unreacted ego vehicle (Section~\ref{sec:exp:validity}). The edit is
applied to the final sample of the diffusion model. We first guided inside the sampler, but the next
denoising step pulled the edited agents back toward nominal driving. This agrees with the observation that denoising pulls samples toward the data manifold~\cite{yang2024diffusiones,bansal2024universal} and that a
constraint defined on the final sample is better enforced there than on the noisy
iterates~\cite{giaretta2026direct}.
For agent $i$ of role $q$,
the relaxed objective is
\begin{equation}\label{eq:relaxed}
\begin{aligned}
J_i(\bu_i)=\frac{1}{\Lambda}\Bigl(&w\,d\bigl(F(s_i(0),\bu_i),\xseed_i\bigr)\\
&+\sum_{k\in\Phi_q^{\mathrm{opt}}}\lambda_k\,\ell\bigl(\rt(\varphi_k,\bx^{(i)},0)\bigr)\Bigr),
\end{aligned}
\end{equation}
with the proximity weight $w=0.05$, the clause multipliers $\lambda_k$, and
$\Lambda=\max(1,\max_k\lambda_k)$. In $\bx^{(i)}$, the staged view of $q$, only agent $i$ is free,
and the other agents of $q$ are held constant by the stop-gradient of~\eqref{eq:detach}. The actions
are updated by projected gradient descent,
\begin{equation}\label{eq:pgd}
\bu_i\leftarrow\Pi_{\Uset}\bigl(\bu_i-\eta\,m_i\,D\,\nabla_{\bu_i}J_i\bigr),
\end{equation}
with a base step size $\eta$ decaying from 0.5 to 0.05 over the 300 steps of a round, the agent's step
multiplier $m_i$ (Section~\ref{sec:method:efficiency}), and $D=\mathrm{diag}(1,0.1)$ for acceleration and
yaw rate. The projection runs between steps, never inside the rollout, sequentially in the speed, and
then bounds the yaw rate by the yaw-rate, lateral-acceleration and curvature limits at that speed.
Algorithm~\ref{alg:staged} summarizes the role-structured solve. Agents present over only part of the horizon keep their
seed.

\subsubsection{Onset and Pinning}
After the attacker's stage, the onset $t^\star$ is computed on the attacker's solution with
the ego vehicle at its seed. If there is no onset, the attacker has not attacked, and the result is
rejected. During the ego vehicle's stage, its actions before the release $\trel=t^\star+\delta$ are held
at $\useed_{\ego}$ (pinning), and only the released actions are optimized. The generated ego vehicle is thus the kind of trajectory that Definition~\ref{def:avoidable} asks for. A hold time of the
attacker, such as the \SI{0.8}{s} of Table~\ref{tab:eb}, is enforced in the same way.

\begin{algorithm}[t]
\caption{Role-Structured Solve\label{alg:staged}}
\begin{algorithmic}[1]
\REQUIRE compiled specification $\Phi$, seed $\xseed$ with actions $\useed$
\STATE $\bu\leftarrow\useed$
\FOR{each role $q$ in priority order}
\IF{$q=\ego$}
\STATE $t^\star\leftarrow$ onset of $\psiobs$ on the staged view of $\att$
\IF{there is no onset} \RETURN rejected \ENDIF
\STATE pin $u_{\ego}(t)=\useed_{\ego}(t)$ for $t<t^\star+\delta$
\ENDIF
\STATE $\lambda_k\leftarrow1$ for $k\in\Phi_q^{\mathrm{opt}}$
\FOR{$n=1,\dots,N_{\mathrm{g}}$}
\STATE run~\eqref{eq:pgd} with step acceptance for all $i\in\Agents_q$
\STATE $\mathcal{F}\leftarrow\{k\in\Phi_q:\rho(\varphi_k,\bx^{(q)},0)<0\}$
\IF{$\mathcal{F}=\emptyset$} \STATE \textbf{break} \ENDIF
\STATE $\lambda_k\leftarrow\gamma\lambda_k$ for $k\in\mathcal{F}\cap\Phi_q^{\mathrm{opt}}$
\ENDFOR
\IF{$\mathcal{F}\neq\emptyset$} \STATE restart from primitives \ENDIF
\ENDFOR
\RETURN $\bx$ and the monitor's verdict on $\bx$
\end{algorithmic}
\end{algorithm}

\subsection{EFFICIENCY MECHANISMS}\label{sec:method:efficiency}\label{sec:method:steps}\label{sec:method:gate}\label{sec:method:bestofk}

Efficiency, as Section~\ref{sec:intro} defines it, is the number of valid scenarios per compute hour and
per recorded input. One run of the solver costs the gradient steps of each role, further rounds with
raised multipliers and restarts when the monitor rejects, and this cost is paid for each of the $K$ seeds
of a window. Four mechanisms reduce it, and none of them changes what the specification requires. A step
rule keeps the gradient steps stable with one step size for all families. The monitor, used as a gate, stops
accepted roles early and restarts rejected ones from maneuver primitives. A pre-check skips seeds that
cannot satisfy the specification. Finally, a batched solver solves the seeds of many windows in one computation on
an accelerator, which is the third contribution. Section~\ref{sec:exp:efficiency} measures their effect.

\subsubsection{Per-Agent Step Acceptance}
A fixed step diverges at high speed, where the curvature of a position
with respect to the yaw rate grows with the square of the speed. Each agent therefore keeps its own step
multiplier $m_i\in[2^{-6},1]$ in~\eqref{eq:pgd}. A step that raises the agent's owned objective by more
than a relative $10^{-3}$ is rejected and halves $m_i$, and ten accepted steps in a row double it, up to
one. The verdict uses only the terms the agent owns, whose gradients it follows, and reuses the value the
next iteration computes anyway. The rule thus keeps the step stable across speed ranges without a step
size per family. Agent-wise gradient normalization~\cite{hetstl2026} and CTG's step
clipping~\cite{zhong2023ctg} perform as well (Section~\ref{sec:experiments}).

\subsubsection{Monitor Gate and Restarts}
After each round of 300 steps, the monitor checks the role's clauses
exactly. An accepted role stops, which saves the remaining rounds. For rejected clauses, the multipliers
are multiplied by $\gamma=4$, for at most $N_{\mathrm{g}}=4$ rounds. A role still rejected restarts from
the best of 585 maneuver primitives (two-phase acceleration offsets combined with lateral shifts of
zero, half or one lane width), scored in one batch by~\eqref{eq:relaxed}. It then restarts from route
primitives
that follow a lane the role's clauses keep it on, which start a vehicle that waits to turn. The monitor
keeps the first accepted result, otherwise the one with the fewest rejected clauses.

\subsubsection{Filtering and Ranking of Candidate Seeds}
The seed generator supplies $K$ candidate seeds per window. On
nuScenes, CTG++ draws them in one batched call, and CTG and CTG++ likewise select among several
samples~\cite{zhong2023ctg,zhong2023ctgpp}. Each candidate first passes a pre-check: the ego's seed is
admissible (Definition~\ref{def:admissible}), the attacker still satisfies its family's attacker selection rule,
and necessary conditions derived from the family's clauses hold. The first sample is always solved. Of
the other remaining candidates, the $n_{\mathrm{s}}-1$ best by a score are solved, where the score is the attack
clauses' hinge loss on the seed itself, one evaluation without optimization. The results are ranked by
acceptance, then by the number of rejected clauses, then by the edit size. With the first sample always
included, the result for $K>1$ is never ranked below the result for $K=1$.

\subsubsection{Batched Solving}
Since the objective is a fixed tensor program for a given specification, the seeds
of a window, and of many windows, are solved in one computation on an accelerator. On a CPU, the batched solver reproduces the sequential one bit for bit in 328 of 329 samples.
Efficient evaluation of STL robustness along the time axis is the subject of~\cite{stlcgpp2025}. Our
batching runs along a different axis, over seed windows and seeds, each judged, gated and accepted
separately. Solving all seeds, $n_{\mathrm{s}}=K$, is then cheap on the accelerator, and each distinct
accepted seed is a
further scenario from the same recorded input.

\section{EXPERIMENTS}\label{sec:experiments}

The experiments follow the three capabilities of Section~\ref{sec:intro}. Generalization is tested by comparing
one parameter set with baselines that need weights, by comparing the specification with loss code as the
output of an LLM, and by counting the NHTSA pre-crash types that the whole pipeline covers
(Section~\ref{sec:exp:generalization}). For validity, we remove the ownership and the stage order of the
role-structured solve one at a time, with everything else unchanged, and check the monitor as the
acceptance gate against the independent judges (Section~\ref{sec:exp:validity}). Section~\ref{sec:exp:efficiency}
measures the effect of batching and of the number of seeds. Finally, we test learned planners
on the generated scenarios and fine-tune one on them (Section~\ref{sec:exp:uses}).

\subsection{SETUP}\label{sec:exp:setup}

\subsubsection{Platforms and Seeds}
On the nuScenes validation split~\cite{caesar2020nuscenes}, which CTG++ was not trained on, the seeds are
samples of the public scene-level CTG++ checkpoint~\cite{zhong2023ctgpp}. Scenes are cut into seed windows
of \SI{5}{s} at a step length of \SI{0.1}{s}. Each vehicle that moves at \SI{3}{m/s} or more and
starts within \SI{1}{m} of a lane center line is bound to the ego role in turn. Unless stated otherwise,
CTG++ draws $K=5$ samples per window, and the method solves the first and the best-scored other one
($n_{\mathrm{s}}=2$).
On CommonRoad, the ego's seed is the plan of Diffusion Planner~\cite{zheng2025diffusionplanner}, a diffusion-based
motion planner with its published nuPlan weights, queried once at the start of the window and cut to the horizon. All
other agents follow their recorded trajectories. A window is dropped when this seed collides, hits a recorded
obstacle or leaves the road before any edit, which happened for 37\,\% of the full set of windows and 54\,\% of the windows
of the baseline comparison. No recorded ego is substituted. The training
experiment and the coverage experiment (Sections~\ref{sec:exp:rq6} and~\ref{sec:exp:rq7}) use the plan of a
sampling-based planner~\cite{werling2010optimal} as the ego seed on CommonRoad. On the same windows, both seeds give the same rate of valid scenarios (Section~\ref{sec:discussion}). The data are the CommonRoad
corpus~\cite{althoff2017commonroad}, the full NGSIM US-101 and I-80 recordings~\cite{ngsim2016}, which
we converted into 390 scenarios with 11,454 cleaned tracks, and the CommonRoad conversion of
nuPlan~\cite{caesar2021nuplan,zhang2023commonroadnuplan} at a step length of \SI{0.2}{s}.
All results use the final version of the method, fixed before the cases were drawn. The runs used a workstation with a 24-core AMD Ryzen
Threadripper 2970WX and two accelerator cards with \SI{11}{GB} of memory each. On the CPU, the solver and
the post-hoc baselines run with one thread per attempt, and CTG++, the batched solver and the learned
planners run on the accelerators. Because the machine was shared with
other jobs, only large ratios of compute times are reported.

\subsubsection{Cases, Judges, and Metrics}
All methods read the same case lists. A seed window is used only if its ego seed is admissible
(Definition~\ref{def:admissible}). Attackers are bound by fixed geometric attacker selection rules per family, for example
the lead vehicle in the ego's lane that the unreacted ego can still reach for lead-vehicle braking (EB). A seed
window with its bound attacker and family is a case, and the cases of a family on a platform form its case set. An
attempt is the run of one method on one case under the budget stated below, so that the attempts of different
methods on a case are paired. The
cases of the baseline comparison were drawn from these case sets by a fixed deterministic rule after the final version
of the method was fixed. Per platform and family, one case from each of ten scenes or map groups forms a development split
for the weighted baseline, and 100 reported cases are drawn round-robin over the remaining groups.
The monitor is the method's own check and never the only criterion of success. Validity is decided by
independent judges that share no code with the specification: the threat (for the cut-in, the unreacted
ego is inside the RSS distance for a braking of \SI{8}{m/s^2}, and for EB it collides with the braking
lead vehicle), an attack onset that the judge determines independently of the observation clause, avoidability
(the generated ego collides with no agent, except when struck from behind), the attacker on the road and clear
of third parties, and on CommonRoad the drivability checker~\cite{pek2020drivability}. A \emph{valid
scenario} passes all of them, and its seed does not. Whether the generated ego reacts before the onset is
reported separately. It decides the pinning ablation and is not part of validity. The judges share the intent of the
specification, but not its code, its onset, its contact test or, on CommonRoad, its road model. The exception for being struck from behind is
used by 14 of the 284 valid scenarios of our method in Table~\ref{tab:rq2main}. Because the generated ego
trajectory is optimized with the attacker's trajectory known, we also check avoidability without that
knowledge. With the attacker fixed, an ego vehicle that keeps its seed until the judge's release, its onset plus
$\delta$, and then brakes in its lane avoids a collision in all 284 valid scenarios of our method in
Table~\ref{tab:rq2main}. The largest deceleration this needs is \SI{2.0}{m/s^2}, and on these scenarios the
judge's onset coincides with the specification's. How hard the ego vehicle has to brake is set by a constant of the
threat clause (Section~\ref{sec:exp:severity}).
Rates come with 95\,\% intervals clustered by nuScenes scene, CommonRoad map group, NGSIM period
window or nuPlan log, written in parentheses, and paired differences with a cluster bootstrap of 10,000
resamples, written in brackets. No correction for multiple comparisons is applied. Pedestrians and
cyclists are modeled with the same unicycle, with the acceleration bounded at \SI{2}{m/s^2} and the speed at
\SI{3.5}{m/s} for pedestrians and at \SI{3}{m/s^2} and \SI{12}{m/s} for cyclists, without a curvature
bound. The full specifications of the four main families are given in Appendix~\ref{app:specs}.

\subsubsection{Baselines}
All baselines edit the same unicycle actions, are projected onto the same bounds, carry the same
proximity term and run four rounds of 300 steps, the budget of one run of our solver per role. They optimize one objective over the
whole horizon without the monitor gate, the multiplier updates and the restarts from primitives, which
are derived from the clause structure. The restarts account for $+0.01$ of our pooled rate, measured on an
earlier run of the same cases with the sampling-planner seed of Section~\ref{sec:exp:setup}. HW (hand-written) is a weighted sum of attack, safety and regularization terms, in the
style of the guidance objectives of~\cite{zhong2023ctg}. B1 optimizes the clauses of our specification as a flat weighted sum in which each
clause reaches all agents it reads, with CTG's step~\cite{zhong2023ctg}. B3 is loss code written by an
LLM in the prompt structure of CTG++~\cite{zhong2023ctgpp}, by DeepSeek V4-Pro with two few-shot examples, ten
drafts per family and no repair loop. B5 pools the robustness of all clauses and
agents into one soft minimum, in the spirit of STRELGen, which pools robustness over all agents~\cite{strelgen2026}. These four optimize
the ego together with the attacker in one objective, as guided diffusion generators that generate all
agents do. For each, X+pin adds our pinning of the ego vehicle before the release, and we report the better
of the two variants. HW
is the only baseline with weights, which we tuned per family on a development split, 480 runs in total.
HW+pin, retuned, repeats the tuning with the pinning on. Our method uses one parameter set
(Section~\ref{sec:method}), which was set during development on pilot sets of a few dozen US-101 and
nuScenes windows per family by inspection of the failures, not by a search. These windows belong to the
same case sets, and the reported cases were drawn afterwards by a fixed deterministic rule. Of the 400, 24 coincide with a pilot case.
We also ran the published guidance of CTG++~\cite{zhong2023ctgpp} on the nuScenes cases, with its collision and
stopping losses, weights and step size as published. These guide all agents of the scene, the ego vehicle included,
inside the denoising. Its plans are fitted to the same unicycle actions before the judges. B0 denotes the unedited
seed of each case.

\subsection{GENERALIZATION}\label{sec:exp:generalization}

\subsubsection{One Parameter Set Against Weighted Objectives}
\begin{table*}[t]
\caption{Rate of Valid Scenarios per Platform and Family, and Paired Differences\label{tab:rq2main}}
\centering\footnotesize\setlength{\tabcolsep}{5pt}
\begin{tabular}{@{}lrrrrrl@{}}
\toprule
Method & CR cut-in & CR EB & nuSc.\ cut-in & nuSc.\ EB & Pooled & Ours $-$ method, pooled \\
\midrule
B0 (seed) & 0.01 & 0.00 & 0.00 & 0.00 & 0.00 & $+0.71$ [$+0.66$, $+0.76$] \\
\midrule
HW & 0.38 & 0.62 & 0.63 & 0.67 & 0.57 & $+0.14$ [$+0.09$, $+0.18$] \\
HW, gate multipliers & 0.31 & 0.54 & 0.68 & \textbf{0.74} & 0.57 & $+0.14$ [$+0.09$, $+0.20$] \\
HW+pin & 0.23 & 0.33 & 0.52 & 0.33 & 0.35 & $+0.36$ [$+0.30$, $+0.41$] \\
HW+pin, retuned & 0.27 & 0.41 & 0.54 & 0.71 & 0.48 & $+0.23$ [$+0.17$, $+0.28$] \\
\midrule
B1 & 0.08 & 0.05 & 0.13 & 0.03 & 0.07 & $+0.64$ [$+0.58$, $+0.69$] \\
B1+pin & 0.08 & 0.05 & 0.18 & 0.02 & 0.08 & $+0.63$ [$+0.58$, $+0.68$] \\
\midrule
B3 & 0.05 & 0.08 & 0.11 & 0.05 & 0.07 & $+0.64$ [$+0.58$, $+0.69$] \\
B3+pin & 0.06 & 0.10 & 0.10 & 0.09 & 0.09 & $+0.62$ [$+0.57$, $+0.68$] \\
\midrule
B5 & 0.21 & 0.06 & 0.29 & 0.15 & 0.18 & $+0.53$ [$+0.48$, $+0.59$] \\
B5+pin & 0.07 & 0.08 & 0.18 & 0.13 & 0.12 & $+0.59$ [$+0.54$, $+0.65$] \\
\midrule
Ours & \textbf{0.52} & \textbf{0.90} & \textbf{0.69} & 0.73 & \textbf{0.71} &  \\
Ours without pin & \textbf{0.52} & \textbf{0.90} & \textbf{0.69} & 0.73 & \textbf{0.71} & $+0.00$ [$+0.00$, $+0.00$] \\
Ours without ownership & 0.04 & 0.00 & 0.00 & 0.00 & 0.01 & $+0.70$ [$+0.65$, $+0.75$] \\
Ours, one stage & 0.43 & 0.85 & 0.65 & 0.70 & 0.66 & $+0.05$ [$+0.03$, $+0.08$] \\
\bottomrule
\end{tabular}
\par\smallskip\parbox{\textwidth}{\footnotesize Valid: passes every independent judge item (on CommonRoad also the drivability checker). $n=100$ attempts per cell, the same cases and one seed for every method; pooled: the 400. +pin: the ego vehicle pinned to its seed until the release. HW: the hand-written weighted sum; HW, gate multipliers: HW with the multiplier schedule of our gate; HW+pin, retuned: weights retuned per family with the pinning on. Ours, one stage: attacker and ego vehicle solved together. Last column: paired difference with a 95\,\% cluster-bootstrap interval. Bold: highest rate per column.}
\end{table*}

Table~\ref{tab:rq2main} compares the methods on the cut-in and EB families, with 100 attempts per platform
and family and one seed each. At $n=100$, a per-cell rate has an interval half-width of about 0.1. Our
method has the highest pooled rate and the highest rate in all columns but one, where HW with the gate multipliers is within 0.01 of it. It leads the strongest baseline, the hand-written
sum HW, by $+0.14$ $[+0.09,+0.18]$ pooled, the retuned HW+pin by $+0.23$ $[+0.17,+0.28]$, and B1, B3 and B5
in their better variant by $+0.53$ to $+0.63$, at one solver run per case. The published guidance of CTG++, in its better
variant per family, yields valid scenarios on 1 of the 100 nuScenes cut-in cases and on 27 of the 100 EB cases, and its
ego vehicle reacts before the onset in all 500 outputs of its five runs (three guidance settings on the braking
cases, two on the cut-in cases), since it guides all agents. With the
best of $K=5$ seeds for each method, our method leads HW on nuScenes by $+0.09$ $[+0.03,+0.14]$ with half as many solver runs (Section~\ref{sec:exp:efficiency}).
Given to HW, so that the weight of each term
not satisfied after a round is multiplied by four, the gate's multiplier schedule leaves the pooled rate at 0.57 (HW, gate multipliers;
$-0.01$ $[-0.05,+0.04]$ against HW). On nuScenes it brings HW to within 0.01 of our rate on both families; on
CommonRoad it lowers HW's rate by $0.07$ and $0.08$. There, the drivability checker catches the
side effects of the terms whose weights were raised: more collisions with third parties, more corridor violations and an ego that
departs from its seed two to five times as far as HW's. A weighted sum has no priority between the attack and
the response. After each round, the headway term stays unsatisfied in 70 to 100\,\% of the EB attempts,
since an EB attack violates the headway by construction, and the collision term in 44 to 74\,\% of the
cut-ins. The schedule therefore raises both sides together, whereas our gate raises a clause's multiplier inside the role that owns it,
after the higher role is fixed. Ours without pin has the same rates, because the attacker is
solved before the ego vehicle and validity does not score the ego's reaction before the onset
(Section~\ref{sec:exp:pinning}). Retuning the weights with the pinning on did not reach the
rate of plain HW. The retuned hand-written sum matches our method only on nuScenes EB ($+0.02$
$[-0.07,+0.11]$), where it needed a 25 times larger attacker weight. This weight moves the attacker by
\SI{28.9}{m} on average against \SI{14.5}{m} for our method (means; Table~\ref{tab:severity} reports medians). The weight search of the retuned HW+pin took 480 runs, 45\,\% of its total compute, and has to be repeated
for each new family or platform.
Over the full case sets of both platforms, lead-vehicle braking succeeds in 0.56 to 0.80 of the attempts and
the cut-in in 0.24 to 0.57, with the same parameters. The nuScenes case sets are run with $K=5$ seeds and two
solver runs per window (Section~\ref{sec:exp:setup}), the CommonRoad case sets with one seed. For the
turning family, the rate is 0.13 on nuScenes and 0.00 to 0.02 on CommonRoad with these parameters, and for
the crossing family 0.15 on
nuScenes (Section~\ref{sec:discussion}). The agent-wise gradient
normalization of~\cite{hetstl2026} and CTG's step clipping perform as well as our per-agent step
acceptance at the default scale.

\subsubsection{Edit Size and Realism}\label{sec:exp:realism}
\begin{table*}[t]
\caption{Edit Size and Departure From Logged Driving of the Valid Scenarios\label{tab:realism}}
\centering\footnotesize\setlength{\tabcolsep}{4.5pt}
\begin{tabular}{@{}lrrrrrrrrrr@{}}
\toprule
& \multicolumn{5}{c}{nuScenes} & \multicolumn{5}{c}{CommonRoad} \\
\cmidrule(lr){2-6}\cmidrule(lr){7-11}
& Edit & \multicolumn{2}{c}{Attacker $W_1$} & \multicolumn{2}{c}{Ego $W_1$} & Edit & \multicolumn{2}{c}{Attacker $W_1$} & \multicolumn{2}{c}{Ego $W_1$} \\
\cmidrule(lr){3-4}\cmidrule(lr){5-6}\cmidrule(lr){8-9}\cmidrule(lr){10-11}
Method & [m] $\downarrow$ & acc.\ $\downarrow$ & jerk $\downarrow$ & acc.\ $\downarrow$ & jerk $\downarrow$ & [m] $\downarrow$ & acc.\ $\downarrow$ & jerk $\downarrow$ & acc.\ $\downarrow$ & jerk $\downarrow$ \\
\midrule
Logs, half against half$^{a}$ &  & 0.15 & 0.43 & 0.15 & 0.43 &  & 0.10 & 0.13 & 0.10 & 0.13 \\
Seed (B0)$^{b}$ & 0.0 & 0.47 & 1.51 & 0.50 & 1.60 & 0.0 & 0.04 & 0.33 & 0.35 & 1.06 \\
\midrule
HW & 8.6 & 0.86 & 1.50 & 0.90 & 1.86 & 8.4 & 1.14 & 0.66 & 1.19 & 1.14 \\
HW+pin, retuned & 10.7 & 0.99 & 1.38 & 0.80 & 1.33 & 10.3 & 1.31 & 1.02 & 1.40 & 0.72 \\
B1+pin & 3.0 & 0.48 & 1.24 & 0.43 & 1.16 & 4.0 & 0.46 & 0.55 & 1.03 & 0.37 \\
B3+pin & 4.4 & 0.88 & 0.88 & 0.19 & 0.53 & 3.7 & 1.18 & 2.04 & 0.20 & 0.37 \\
B5 & 7.7 & 0.87 & 1.63 & 1.28 & 0.95 & 9.2 & 1.01 & 1.12 & 1.97 & 0.58 \\
\midrule
Ours & 4.8 & 0.89 & 0.83 & 0.57 & 1.64 & 5.8 & 1.08 & 0.68 & 0.71 & 0.72 \\
Ours, one stage & 7.6 & 1.33 & 0.88 & 1.05 & 1.37 & 7.9 & 1.56 & 0.54 & 1.44 & 0.50 \\
\bottomrule
\end{tabular}
\par\smallskip\parbox{\textwidth}{\footnotesize Valid outputs of the 200 attempts per platform of Table~\ref{tab:rq2main}. Edit: mean distance of the attacker from its seed over the horizon. $W_1$: Wasserstein-1 distance between the pooled \SI{2}{Hz} acceleration [\si{m/s^2}] and jerk [\si{m/s^3}] of the generated vehicles and those of the logged vehicles of the same scenes; the attacker over the \SI{5}{s} horizon, the ego vehicle from the judge's attack onset on. $^{a}$Two random halves of the logged vehicles against each other, mean of 200 splits. $^{b}$The seed attacker and ego vehicle over the whole horizon of all attempts.}
\end{table*}

Table~\ref{tab:realism} reports how far each method moves the attacker and how far the kinematics of its
valid scenarios depart from logged driving, measured as the Wasserstein-1 distance between the pooled
acceleration and jerk of the generated vehicles and those of the logged vehicles of the same scenes. For the attacker, the distance is measured over the horizon, and for the ego vehicle from the judge's onset on. The cells hold the valid
outputs of each method, 13 to 142 per platform. The differences below are paired over the cases valid for both
methods; Table~\ref{tab:realism} lists the means over each method's own valid outputs. Since all methods edit the same unicycle actions under the
same acceleration and yaw-rate bounds, the yaw-rate distances (0.01 to 0.10 for each method) are omitted.
Two references set the scale: two halves of the logs differ by 0.15 and 0.10 in acceleration on the two platforms, and the seed
attackers by 0.47 and 0.04. Since an attack is not typical driving, the attacker of each method departs further
than its seed, ours by 0.89 and 1.08. In this respect our method does not differ from the joint objectives HW,
HW+pin and B5: the paired differences in the attacker's acceleration lie within $\pm0.1$, and all intervals include zero. Their attackers, however, are moved 2.8
to \SI{5.7}{m} further from the seed, and their jerk departs
further on nuScenes, by 0.53 to 0.60. The response differs. After the onset, the ego vehicles of HW, HW+pin
and B5 depart further from the logs than ours by 0.23 to 1.16 in acceleration, with all intervals above zero. Ours departs by 0.57 and 0.71, against 0.52 and 0.44 for the unreacted seed ego measured over the
same windows (the seed row of Table~\ref{tab:realism} is measured over all windows). Solving both roles in one
stage moves the attacker 2.3 to \SI{2.7}{m} further and raises the departure in acceleration of the attacker
by 0.41 and 0.50 and of the ego vehicle by 0.46 and 0.74. B1+pin and B3+pin, valid on 7 to 10\,\% of the
attempts, stay as close to the logs in acceleration as our method. On nuScenes, the
jerk of the CTG++ seed itself is 1.5 from the logs against a floor of 0.43. A jerk distance below the
seed's therefore means that the solver smoothed the sample.

\subsubsection{Severity as a Specification Constant}\label{sec:exp:severity}
\begin{table}[t]
\caption{Severity as a Specification Constant: Lead-Vehicle Braking on nuScenes, the 100 Cases of Table~\ref{tab:rq2main}\label{tab:severity}}
\centering\scriptsize\setlength{\tabcolsep}{3pt}
\begin{tabular}{@{}lrrrrr@{}}
\toprule
& & Valid and & \multicolumn{2}{c}{Required deceleration [m/s$^2$]} & Attacker \\
\cmidrule(lr){4-5}
Constant $a$ & Valid & accepted & median [Q1, Q3] & $\geq 2$ & edit [m] \\
\midrule
None (Table~\ref{tab:rq2main}) & 73 & 46 & 0.50 [0.25, 0.75] & 1/73 & 12.1 \\
2 & 72 & 46 & 1.50 [0.75, 2.00] & 24/72 & 19.1 \\
3 & 68 & 38 & 1.75 [1.00, 2.31] & 32/68 & 19.1 \\
4 & 64 & 33 & 2.00 [1.25, 2.75] & 33/64 & 19.1 \\
\bottomrule
\end{tabular}
\par\smallskip\parbox{\columnwidth}{\footnotesize Valid: passes every independent judge item; accepted: the monitor accepts. Required deceleration: the in-lane braking from the release that the unreacted ego vehicle needs to avoid contact within the horizon, over the valid outputs. Edit: median over the valid outputs of the attacker's largest displacement from its seed.}
\end{table}

The threat clause of a specification states how dangerous a scenario has to be, and the solver returns the smallest
edit of the seed that meets it. Under the lead-vehicle braking specification (Table~\ref{tab:spec:eb}) as run for Table~\ref{tab:rq2main} the threat is a
collision with the unreacted ego vehicle, and the solver meets it with the lightest braking that produces one, late in the
horizon. An ego vehicle that reacts at the release avoids it with a median deceleration of \SI{0.5}{m/s^2}
(Table~\ref{tab:severity}, first row). A harder scenario is asked for by a constant rather than by a weight. We add to the
threat clause that the lead vehicle brakes until it stops and that at the release the unreacted ego vehicle already needs
more than $a$ to stop behind it. This is the required-deceleration criterion of the cut-in threat, placed
at the release. We ran the same solver with the same budget and judge on the 100 nuScenes
lead-vehicle braking cases of
Table~\ref{tab:rq2main} for $a$ of 2, 3 and \SI{4}{m/s^2}. The median deceleration that the valid scenarios require
rises from 0.5 to \SI{2.0}{m/s^2}, the number of scenarios that require at least \SI{2}{m/s^2} from 1 of 73 to 33 of
64, and the hardest scenario requires \SI{4.0}{m/s^2}, next to the ego vehicle's own bound of \SI{4.5}{m/s^2}. From
$a=3$ on, the scenarios require more than those of HW and of the retuned HW+pin on the same cases, whose medians are
1.0 and \SI{1.5}{m/s^2}. At $a=3$, the rate of valid scenarios, 0.68, matches that of HW at its tuned weights, 0.67,
and the scenarios require 1.75 against \SI{1.0}{m/s^2}. A threshold placed in a term of HW would not enter its
acceptance, since HW has none, and the term's weight would have to be searched again. The
ego vehicle's response grows from 1.3 to \SI{9.3}{m} of displacement and the attacker's edit from 12 to \SI{19}{m}.
The price is validity: the rate falls from 0.73 to 0.64, and the gate accepts 33 scenarios instead of 46. Both fall because the
ego vehicle's braking bound and its delay of \SI{0.6}{s} leave a narrowing band of avoidable scenarios
and the gate rejects the solver runs that fall short of the constant, most often on the attack clause. The
measured deceleration stays below the constant, which bounds a stopping distance with a floor of \SI{2}{m} whereas
the measurement is the braking that avoids the collision within the horizon. Such a criterion is already the threat of the
cut-in, with the friction limit of \SI{8}{m/s^2} as its constant (Appendix~\ref{app:specs}). For the
cut-in, we placed the same constant after the lane entry, so that the attacker brakes until it stops once
in the ego vehicle's lane and the unreacted ego vehicle needs more than $a$ at the release. At $a=4$, this
gives valid scenarios on 62 of the 100 nuScenes cut-in cases of
Table~\ref{tab:rq2main}, 46 of them accepted, against 69 and 56 under the default clause.

\subsubsection{LLM-Drafted Specification Versus Loss Code}\label{sec:exp:rq1}
\begin{table}[t]
\caption{Valid Scenarios From LLM-Drafted Specifications and Loss Code\label{tab:rq1}}
\centering\footnotesize\setlength{\tabcolsep}{3pt}
\begin{tabular}{@{}lrrl@{}}
\toprule
LLM & STL & Code & STL $-$ code \\
\midrule
Claude Opus 5.5 & 9/34 & 5/34 & $+0.12$ [$+0.00$, $+0.31$] \\
GPT-6.1 Sol & 14/42 & 8/42 & $+0.14$ [$-0.05$, $+0.31$] \\
DeepSeek V4-Pro & 32/168 & 16/168 & $+0.10$ [$-0.01$, $+0.22$] \\
Claude Sonnet 5.5 & 10/34 & 6/34 & $+0.12$ [$-0.09$, $+0.38$] \\
GPT-6 Luna & 64/252 & 39/252 & $+0.10$ [$+0.00$, $+0.20$] \\
DeepSeek V4.1-Flash & 23/168 & 26/168 & $-0.02$ [$-0.12$, $+0.11$] \\
\midrule
All & 152/698 & 100/698 & $+0.08$ [$-0.00$, $+0.19$] \\
\bottomrule
\end{tabular}
\par\smallskip\parbox{\columnwidth}{\footnotesize Valid attempts over all descriptions and windows on nuScenes. Code: loss code optimized with our step rule and pinning. STL $-$ code: paired difference with a 95\,\% bootstrap interval over NHTSA types; All: cells weighted equally.}
\end{table}

A new scenario type is stated by a description in natural language. We compare the specification with loss
code as the output of six LLMs from three providers. The descriptions are those of the typical scenarios
of the NHTSA pre-crash types~\cite{nhtsa2007typology}, with three paraphrases each, on the 14 types that
ran on nuScenes at the time. Later, the two cyclist types of Table~\ref{tab:rq7} were added. The code variant follows CTG++~\cite{zhong2023ctgpp} with few-shot examples of the same
clauses and shares the optimizer, the step rule, the pinning with its own onset, the budget and the repair
loop. Only the representation and what it compiles into differ between the two variants. The specification variant runs the
role-structured solve of Section~\ref{sec:method:staged}, and the code variant optimizes its code as one
joint objective, as CTG++ does. We therefore compare the two pipelines and not the two languages alone. An attempt is valid if it passes the generic judge and a judge of the requested
type, and its seed does not. The per-type judges were written by us from the NHTSA definitions before the
runs, independently of the specifications. Each LLM was queried through OpenRouter, restricted to its
first-party provider, in reasoning mode at the provider's default sampling settings, in October 2026. The
model identifiers are listed in Appendix~\ref{app:specs}.
The difference fixed before the runs, on the official descriptions of the 12 types that all LLMs ran, with the LLMs
weighted equally, is $+0.04$ $[-0.08,+0.14]$, an interval that includes zero (Table~\ref{tab:rq1}, whose per-model
rows cover the types each model ran, which differ between models).
Over all descriptions and the 14 types, the specification gives more valid attempts for five of the six LLMs, 152
of 698 against 100 of 698. Per type, the largest difference is for turning at non-signalized junctions, $+0.59$
$[+0.45,+0.73]$. Run with CTG's step instead of ours, loss code yields 124 of 698 valid attempts, still
below the specification's 152. The type check rejected 111 of the 698
first drafts of a specification, 61 of which could not be executed at all. Run unchecked, these drafts gave
3 valid attempts. Checked and repaired, they gave 15, a paired difference of $+10.8$ $[+3.7,+19.6]$
percentage points on the rejected drafts.
The variance comes mostly from the seed window and the draft of the LLM. For several types one window
yields about half valid attempts and the other none, in both variants. Two drafts for the same description and window differ in validity in 52 of the 75 pairs with at least one valid draft. In 408 of the 543 invalid
attempts of the specification variant that reached the solver, a road clause is among the clauses the solver cannot satisfy, typically a
crossing or turning attacker confined to the ego's road corridor from $t=0$. Claude Opus 5.5 returned no
content for 8 queries of the code variant because of its provider's content filter, which we count as refusals.

\subsubsection{Coverage of Pre-Crash Types}\label{sec:exp:rq7}
\begin{figure*}[t]
\centering
\includegraphics[width=\textwidth]{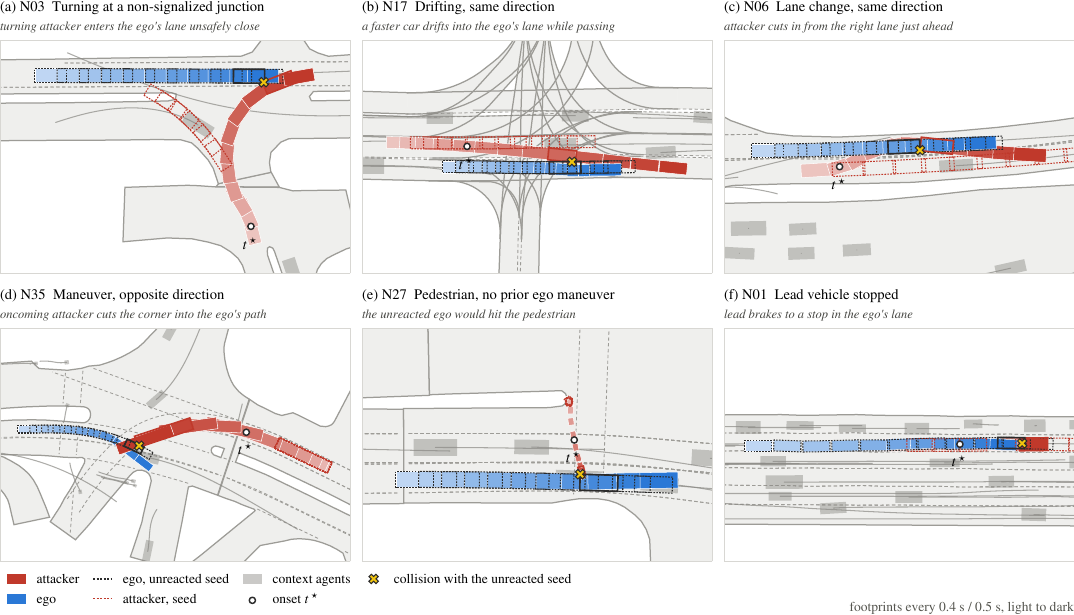}
\caption{Generated scenarios of six NHTSA pre-crash types, each from an LLM-drafted specification, with the parameter
set of all other experiments and no weights: (a), (c)--(e) nuScenes, (b), (f) CommonRoad. The line under each title
paraphrases the attack clause of the draft. In every panel the attacker's seed (dashed red) did something else, the
attack is turned against the unreacted seed of the ego vehicle (dashed black), with which it collides, and the generated ego
vehicle responds from the release on with the smallest edit that keeps it clear of the attacker, by 0.1 to \SI{0.7}{m} at
the closest point: it brakes in (a), (c), (e) and (f) and swerves in (b) and (d). The six pass the independent judges of their
types. The drafted road clauses, which confine the attacker to the ego vehicle's road corridor, are not met at the
junctions, which is why the specification's own monitor accepts only (b), (c) and (f) (Section~\ref{sec:discussion}).
Footprints every \SI{0.5}{s} (\SI{0.4}{s} on CommonRoad), light to dark; context agents in grey.\label{fig:gallery}}
\end{figure*}
We count coverage in the pre-crash typology of the NHTSA~\cite{nhtsa2007typology} because it lists
concrete types together with their shares of crashes, so that coverage can be weighted by crash
frequency. ISO 34502~\cite{iso34502} structures the test scenarios of automated driving systems by the
disturbances to perception, traffic and vehicle motion and prescribes no such list; its traffic scenarios are related
to the types below in Appendix~\ref{app:coverage}. Of the 37 NHTSA pre-crash types, 23 involve another road user and are
applicable to trajectory-level generation. They cover 71.3\,\% of the crashes, and the other 14 involve a
single vehicle, an object, an animal, or an evasive maneuver.
Sixteen of the 23 types can be expressed with the agents and maps of the public data, with 57.8\,\% of the crashes. Two need signal states that the scenes do not contain, and
five need agents or maneuvers that the scene adapters do not provide, such as reversing or a parked
attacker. We ran each of the 16 on windows of both platforms where they exist, with DeepSeek V4-Pro,
four descriptions and two drafts per window. The pipeline of drafting, type check, compilation and generation
yields valid scenarios for 11 of the 16 types, which together account for 46.9\,\% of all crashes, among them
turning at junctions, lane changes, a pedestrian and a cyclist crossing. Six of them are shown in
Fig.~\ref{fig:gallery}, and
Table~\ref{tab:rq7} in Appendix~\ref{app:coverage} gives the attempts, windows and valid scenarios per type. These 16 are
NHTSA types selected by what the public data can express. They are not the 16 interaction modes that
CrashSim derives from the same typology to build a crash prior on nuScenes, where the ego vehicle replays
the log~\cite{peng2026crashsim}.
Appendix~\ref{app:coverage} relates two catalogues written for testing rather than from crash
statistics to these runs (Table~\ref{tab:catalogues}). They are the traffic-related critical scenarios of
ISO 34502, which
structures highway scenarios by road sector, the ego vehicle's behavior and the motion of a surrounding vehicle, and
the AEB scenarios of the Euro NCAP 2023 protocols. Four of the seven Euro NCAP car-to-car scenarios have been generated on recorded data. Two
others, the straight crossing and the oncoming vehicle drifting into the ego vehicle's lane, were
drafted and run without a valid output. For the seventh, the ego vehicle's turn across an oncoming vehicle, no window of
the public data with a turning ego vehicle has an admissible seed. Of the four surrounding-vehicle motions of ISO 34502, the cut-in
and the deceleration of the lead vehicle have been generated. The cut-out and the acceleration of a follower have
a specification that the solver satisfies on a synthetic scene but have not been run on recorded data. For the ISO 34502
scenarios with a lane-changing ego vehicle and the motorcycle scenarios, specifications exist, but they
wait for seed windows or agents that our case sets do not yet provide.

\subsection{VALIDITY}\label{sec:exp:validity}

\subsubsection{Ownership}
When our solver lets each clause's gradient reach both vehicles, with everything else unchanged, the rate of
valid scenarios drops from 0.71 to 0.01, by $0.70$ $[0.65,0.75]$ (Table~\ref{tab:rq2main}). Without ownership, the ego's
stage satisfies its clauses most cheaply by moving the attacker back, which deviates at most 1.9 to \SI{4.2}{m} from
its seed against 14.1 to \SI{19.1}{m} with ownership. The baselines that optimize one objective without
ownership reach the threat in 0.08 to 0.34 of the attempts with the pinning, against 0.70 to 0.95 for our
method. With the multiplier growth of the gate turned off
($\gamma=1$) in both variants, on the same 400 cases, the rate without ownership stays at 0.01, no output passes
the monitor, and the gap to our method is still $0.57$ $[0.52,0.63]$. Our method itself loses $0.13$
$[0.10,0.16]$ without the growth, mostly in the cut-in, whereas the same growth given to the hand-written sum
gives no gain (Section~\ref{sec:exp:generalization}).

\subsubsection{Stage Order}
When the attacker and the ego vehicle are solved together, with ownership, gate, step rule, restarts and
budget unchanged, the attack clauses read the responding ego vehicle. As a result, the attacker chases the response. Its onset comes earlier and it brakes harder. In the cut-in, it collides with the unreacted ego
vehicle in 41\,\% of the attempts on both platforms, against 1\,\% on CommonRoad and 0\,\% on nuScenes with
the stage order, although the cut-in specification asks for an unsafe gap and not a collision. The rate of valid
scenarios falls from 0.71 to 0.66, by $0.05$ $[0.03,0.08]$, and a solver run takes 1.9 times as long. Under the
specification's monitor, acceptance falls from 0.51 to 0.32. Both variants stay avoidable. For this check
(Section~\ref{sec:exp:setup}), the attacker is kept fixed, and the ego vehicle may respond only from the
release $t^\star+\delta$, by our response stage or by braking at up to \SI{8}{m/s^2} in its lane. None of
the accepted scenarios of either variant is unavoidable.

\subsubsection{Acceptance Gate}
The monitor of the specification marks which outputs are valid. Of the outputs it accepted, 203 of 204 in
Table~\ref{tab:rq2main} and 115 of 115 at $K=5$ passed the independent judges. The one exception is a
CommonRoad cut-in that only the drivability checker rejects. The monitor is the
stricter of the two, since it also rejected 81 of the 284 outputs that the judges accepted. In contrast, the baselines
return all outputs without such a mark. Of the outputs of HW, 43\,\% are invalid.

\subsubsection{Pinning}\label{sec:exp:pinning}
Pinning the ego vehicle until the release keeps the generated ego from reacting before the attack is
observable. Without it, the ego deviates from its seed before the onset in 0.87 of the 400 CommonRoad attempts
with the sampling-planner seed (Section~\ref{sec:exp:setup}). Because
the attacker is solved first, pinning does not change the attack. The attack is identical with and
without pinning in all 400 attempts of that split and has the same edit statistics on the 200 CommonRoad
cases of Table~\ref{tab:rq2main}. Pinning also leaves the rate of valid scenarios and the avoidability
after the release unchanged, and it has no measurable effect on fine-tuning (Section~\ref{sec:exp:uses}).

\subsection{EFFICIENCY}\label{sec:exp:efficiency}
\begin{table*}[t]
\caption{Number of Seeds $K$ With All Seeds Solved in Batches on the Accelerator\label{tab:ksweep}}
\centering\footnotesize\setlength{\tabcolsep}{6pt}
\begin{tabular}{@{}rrllrrl@{}}
\toprule
$K$ & Runs & Valid $\uparrow$ & Distinct $\uparrow$ & Edit [m] & Win./h & Ours $-$ HW+pin, ret. \\
\midrule
1 & 1.0 & 0.71 (0.63--0.78) & 0.51 (0.43--0.59) & 14.7 & 998 & $-0.02$ [$-0.07$, $+0.03$] \\
5 & 3.9 & 0.78 (0.69--0.85) & 1.51 (1.26--1.78) & 12.7 & 612 & $+0.04$ [$-0.01$, $+0.09$] \\
\textbf{10} & 7.4 & 0.80 (0.73--0.87) & 2.35 (1.94--2.75) & 11.9 & 467 & $+0.07$ [$+0.03$, $+0.11$] \\
20 & 14.3 & 0.81 (0.73--0.88) & 3.38 (2.82--3.94) & 10.5 & 310 & $+0.08$ [$+0.03$, $+0.13$] \\
\midrule
\multicolumn{7}{@{}l}{$K=5$, windows per batch 1 / 128: 74 / 654 windows per accelerator hour} \\
\bottomrule
\end{tabular}
\par\smallskip\parbox{\textwidth}{\footnotesize 200 nuScenes attempts (100 cut-in, 100 EB) of the baseline split. Runs: solver runs per attempt, one per solved seed. Valid: rate of valid scenarios of the selected result, with a 95\,\% clustered interval. Distinct: valid scenarios that the monitor also accepts and whose attackers differ by at least \SI{1}{m}, per seed window. Edit: mean maximum attacker distance from its seed. Win./h: seed windows solved per hour on one accelerator card. Last column: paired difference against the retuned HW+pin at $K=5$. Bold: the recommended $K$.}
\end{table*}

The batched solver uses the fact that a compiled specification is a fixed tensor program
and stacks the seeds of many windows in one computation on the accelerator. On the CPU, it reproduces the sequential solver bit
for bit in 328 of 329 samples, and the remaining sample differs by \SI{2e-6}{m} with the same verdict. With $K=5$, one
window per batch gives 74 windows per accelerator hour, timed on 16 of the 128 windows, 25 windows per batch give
612 (Table~\ref{tab:ksweep}), and 128 windows in one batch give 654, about nine times as many as one window.
With all seeds solved, a larger $K$ yields more and more diverse valid scenarios per recorded input
(Table~\ref{tab:ksweep}). On 200 nuScenes attempts, the rate of valid scenarios grows from 0.71 at $K=1$ to
0.80 at $K=10$, and the number of distinct valid scenarios per seed window from 0.51 to 2.35. Beyond
$K=10$ the rate no longer grows, 0.81 at $K=20$, while the solver runs per window do. We therefore recommend $K=10$. The
throughput of Table~\ref{tab:ksweep} counts the solver alone, 25 windows per batch on a shared machine, without
CTG++ sampling and the unicycle fit, and the solver runs are matched to the retuned HW+pin only in the row $K=5$.
Pre-scoring loses little: solving two of five seeds gives a rate within 0.01 of solving all five. Of the
3,800 additional samples of the sweep, 30\,\% were dropped before any solver run, a sample often for more than one reason: 592 because the seed was not admissible, 335
because the sample offered no attacker for the family, and 286, all cut-ins, by the pre-check of the attack clauses.
On one CPU thread, a solver run of our method takes about twice as long as one of the retuned HW+pin
(\SI{194}{s} against \SI{91}{s}). The gain in throughput comes from batching. With equal numbers of
solver runs at $K=5$, our method leads the retuned HW+pin by $+0.04$ $[-0.01,+0.09]$. Plain HW, the strongest
baseline at one solver run, gains from the best of five seeds on the cut-in ($+0.08$ $[+0.02,+0.15]$) and not on the EB ($-0.02$ $[-0.08,+0.04]$). The reason is that its selection keeps the seed with the smallest weighted loss, which need
not be the one the judge prefers. Under this rule, HW solves 3.9 seeds per case against 1.9 for ours with pre-scoring,
and our gain over it on nuScenes is $+0.09$ $[+0.03,+0.14]$ at $K=5$ against $+0.06$ $[+0.01,+0.12]$ at $K=1$.
The CTG++ draws and the fit of the extra seeds are not batched and are excluded from these timings.

\subsection{USES: TESTING AND TRAINING PLANNERS}\label{sec:exp:uses}

\subsubsection{Testing Learned Planners}\label{sec:exp:rq5}
\begin{table*}[t]
\caption{Planner Failure Rate on Avoidable Scenarios, All Families\label{tab:rq5}}
\centering\footnotesize\setlength{\tabcolsep}{4pt}
\begin{tabular}{@{}lllllll@{}}
\toprule
Platform & Planner & Ours [\%] & Seed B0 [\%] & Ours $-$ B0 [pp] & HW+pin, retuned [\%] & Ours $-$ HW+pin, retuned [pp] \\
\midrule
nuScenes & Diffusion Planner & 6 (2--10) & 0 (0--1) & $+5$ [$+2$, $+10$] & 11 (6--16), $n=125$ & $-5$ [$-8$, $-2$], $n=89$ \\
nuScenes & Flow Planner & 17 (11--24) & 2 (0--5) & $+15$ [$+9$, $+21$] & 46 (36--57), $n=125$ & $-38$ [$-47$, $-29$], $n=89$ \\
CommonRoad & Diffusion Planner & 13 (8--20) & 3 (0--7) & $+10$ [$+3$, $+18$] & -- & -- \\
CommonRoad & Flow Planner & 18 (8--28) & 4 (1--8) & $+15$ [$+5$, $+25$] & -- & -- \\
CommonRoad & Reactive (reference) & 8 (3--14) & 1 (0--4) & $+7$ [$+2$, $+12$] & -- & -- \\
\bottomrule
\end{tabular}
\par\smallskip\parbox{\textwidth}{\footnotesize Failure: at-fault collision or leaving the drivable area in closed loop, mean over four rollouts per scenario (one for the deterministic reactive planner), then over scenarios, with a 95\,\% cluster-bootstrap interval. Ours and B0: the same 100 scenarios per platform; B0 is the unedited seed of each. HW+pin, retuned: paired on the cases for which both methods produced an eligible scenario.}
\end{table*}

We ran Diffusion Planner~\cite{zheng2025diffusionplanner} and Flow Planner~\cite{tan2025flowplanner}
with their published weights through their own wrappers, re-planning at \SI{10}{Hz}, with four rollouts per
scenario. The reactive planner serves as a reference. A rollout fails on an at-fault collision or on
leaving the drivable area. Per platform, 100 of our valid scenarios, all avoidable by the judges, and
their unedited seeds were run. On CommonRoad the seed is the plan of Diffusion Planner itself, and one case
raised an error for the reactive planner, which ran on 99. The retuned HW+pin was compared on the nuScenes
baseline split: an output is eligible when it passes the judges and has no collision with a third party, which holds
for 125 of its 200 outputs, and the paired difference uses the 89 cases on which both methods produced an
eligible scenario. Its CommonRoad
comparison ran with the sampling-planner seed (Section~\ref{sec:discussion}).
Both learned planners fail more often on our scenarios than on the same seeds, by $+5$ to $+15$
percentage points with all intervals above zero (Table~\ref{tab:rq5}), and almost all added failures are
at-fault collisions with the attacker. The added failures concentrate where the attack crosses the ego
vehicle's path or brakes in front of it. On nuScenes, Diffusion Planner fails $+50$ $[0,+100]$ and $+20$ $[-2,+57]$ points more
often on the 4 turning (LTAP) and 11 crossing (VRU) scenarios and $+2$ $[0,+7]$ on the cut-ins, and Flow Planner $+11$
$[+3,+18]$ more often on lead-vehicle braking. On CommonRoad, both planners add $+12$ $[+4,+22]$ and $+17$ $[+3,+29]$ points on
lead-vehicle braking and $+4$ $[-3,+16]$ and $+9$ $[+1,+18]$ on the cut-in.
No failure on a seed involves the attacker. The reactive planner, run through a different simulation framework, fails
in 8\,\% of our scenarios and 1\,\% of the seeds, $+7$ $[+2,+12]$ percentage points. Seven of its eight
failures are at-fault collisions with the attacker in scenarios that the judges' braking response avoids.
The scenarios of the retuned HW+pin, whose attacker is moved \SI{23.7}{m} against \SI{9.4}{m} for ours (median),
make the learned planners fail more often still on nuScenes (Table~\ref{tab:rq5}). Raising the severity constant of
Section~\ref{sec:exp:severity} to $a=4$ brings our scenarios to that level. On the 64 nuScenes lead-vehicle braking
cases of Table~\ref{tab:rq2main} that are valid at $a=4$, Flow Planner fails on 42\,\% of the scenarios
against 13\,\% of the scenarios generated with the default constant, $+29$ $[+18,+38]$ percentage points
paired on the same cases. Diffusion Planner fails on 11\,\% against 8\,\%, $+3$ $[0,+6]$. The 82
CommonRoad lead-vehicle braking cases of Table~\ref{tab:rq2main} that are valid at $a=4$ make both
planners fail several times as often: Flow Planner on 57\,\%
against 14\,\%, $+43$ $[+32,+52]$, and Diffusion Planner on 28\,\% against 9\,\%, $+20$ $[+11,+29]$, almost always by
an at-fault collision with the braking lead vehicle. For the cut-in, 62 nuScenes cases are valid at $a=4$, with the attacker braking
to a standstill after the lane entry (Section~\ref{sec:exp:severity}). On them, Flow Planner fails on
58\,\% against 9\,\%, $+49$ $[+38,+65]$, and Diffusion Planner on 15\,\% against 2\,\%, $+12$ $[+6,+20]$,
and all failures are at-fault collisions with the braking attacker. These scenarios require the planner to brake once the attack is observable. In 33 of the 64 nuScenes braking cases, the ego vehicle needs
at least \SI{2}{m/s^2} from the release to stop behind the lead vehicle (Table~\ref{tab:severity}).

\subsubsection{Training a Learned Planner}\label{sec:exp:rq6}
We fine-tune Diffusion Planner from its published weights on a replay of its own nuPlan training data,
to which a share of generated scenarios is added, and test it in closed loop. Training and test scenarios
come from CommonRoad and are disjoint by recording period on NGSIM and by map on the CommonRoad corpus.
On the training side, our method generated 1,121 scenarios. We compare them with replay only, with
the same 1,121 seeds unedited, with our scenarios without pinning, and with two variants of the same
scenarios. The first adds one clause on the ego label that keeps a time headway of \SI{1.5}{s} from
\SI{0.2}{s} after the onset (1,113 pass the gate). The constant \SI{1.5}{s} was chosen on the validation split. The second adds that clause together with the severity
constant $a=4$ of Section~\ref{sec:exp:severity} on the
braking scenarios (802 pass the gate). As a control, HW generated its scenarios with the time headway of its
safety term raised from 1.0 to \SI{1.5}{s}, the headway of that clause, and nothing else changed (803 pass the
judges). We chose the share, 1/9 of the training
samples, on a validation split by a rule written before the evaluation. Each training set was used in three
training runs with different random initializations, each evaluated once on the test split.
The test split holds 616 critical scenarios
generated by our method on which the published planner or the replay-only planner fails or passes within
\SI{4.6}{m} of the attacker, 640 critical scenarios generated by the retuned HW+pin on the same windows,
the 559 unedited seeds of the critical scenarios as benign twins, and 60 nuPlan logs. Since the critical
test set is selected on the outcomes of the published and the replay-only planners, it is biased against the
replay-only planner. The comparisons between the planners trained on generated data do not depend on this selection.
\begin{table}[t]
\caption{Fine-Tuning Diffusion Planner on Generated Scenarios\label{tab:rq6}}
\centering\scriptsize\setlength{\tabcolsep}{2pt}
\begin{tabular}{@{}lrrrrrrr@{}}
\toprule
 & & & \multicolumn{2}{c}{Critical [\%] $\downarrow$} & \multicolumn{3}{c}{Benign twins} \\
\cmidrule(lr){4-5}\cmidrule(lr){6-8}
Added data & Scenes & Headway & ours & HW+pin & Fail. & Progress & Hard br. \\
\midrule
None (replay) & -- & -- & 36.8 & 55.3 & 5.4 & -- & 0.8 \\
Unedited seeds & 1,121 & -- & 40.5 & 57.7 & 4.9 & $+0.013$ & 0.5 \\
Ours & 1,121 & 0.84 & 34.0 & 53.4 & 3.3 & $-0.011$ & 0.6 \\
Ours, no pinning & 1,121 & -- & 32.4 & 52.1 & 2.8 & $-0.017$ & 0.6 \\
B1 & 101 & -- & 45.0 & 61.6 & 5.9 & $+0.024$ & 0.9 \\
HW+pin, ret. & 525 & 0.91 & 29.8 & 50.5 & 1.8 & $-0.031$ & 0.4 \\
HW & 871 & 1.25 & 28.2 & 47.4 & 2.5 & $-0.026$ & 0.5 \\
\midrule
Ours, \SI{1.5}{s} & 1,113 & 1.51 & 26.7 & 46.5 & 1.8 & $-0.037$ & 0.5 \\
Ours, \SI{1.5}{s}, $a=4$ & 802 & 1.41 & 24.2 & 43.1 & 1.8 & $-0.042$ & 0.6 \\
HW, \SI{1.5}{s} & 803 & 1.63 & 21.7 & 42.5 & 1.6 & $-0.045$ & 0.8 \\
\bottomrule
\end{tabular}
\par\smallskip\parbox{\columnwidth}{\footnotesize Failure: at-fault collision or leaving the drivable area in closed loop, mean over three training runs. Headway: median over the scenarios of the minimum time headway of the ego label, in s; \SI{1.5}{s}: the label keeps a time headway of \SI{1.5}{s}, by the clause (ours) or the safety term (HW). Fail.\ and Hard br.\ in \%. Critical: 616 test scenarios of our method and 640 of the retuned HW+pin. Benign twins: the 559 unedited seeds of the critical scenarios; Progress: travelled distance over the recorded driver's, as a difference from the replay-only planner; Hard br.: rollouts braking below $-3$\,m/s$^2$.}
\end{table}

Fine-tuning on our scenarios lowered the failure rate on the critical scenarios from 36.8 to 34.0\,\%
(Table~\ref{tab:rq6}), by $-2.8$ $[-4.2,-1.0]$ percentage points, and on those of HW+pin by $-1.9$
$[-3.7,-0.1]$. Since the unedited seeds raised it, the gain comes from the attacks and their responses.
Against the unedited seeds, our scenarios lowered it by $-6.5$ $[-9.4,-4.5]$. On the benign twins, the
progress relative to the replay-only planner changed by $-0.011$ $[-0.019,-0.001]$ and hard braking by $-0.2$ $[-0.7,+0.0]$
percentage points. Scenarios
without pinning gave a similar gain. The 101 valid scenarios of B1 on the same
windows, oversampled to the same share, raised the failure rate, and ours lowered it by $-11.0$ $[-15.0,-8.2]$ against them.

The scenarios of the retuned HW+pin and of HW lowered the failure rate by a further $4.2$
$[2.9,5.6]$ and $5.7$ $[4.0,8.3]$ percentage points. Fine-tuned on them, the planners drive
more cautiously: before the attack they planned 0.13 to \SI{0.17}{m/s} slower, and their progress on the benign
twins fell two to three times as much as with our default label. In the ego trajectories of HW, a safety
term keeps a time headway of \SI{1}{s} from the first step, so its labels teach the planner to keep a distance
before the attack. Two comparisons isolate the time headway of the label. On the same attacks (1,113 of the 1,121
windows pass the gate), the clause that keeps a time headway of \SI{1.5}{s} from \SI{0.2}{s} after the onset lowers the failure
rate from 34.0 to 26.7\,\% on our critical scenarios and from 53.4 to 46.5\,\% on those of HW+pin, by $-7.3$
$[-10.1,-5.3]$ and $-6.9$ $[-11.4,-4.5]$ percentage points, and the planner's progress on the benign twins falls by
a further $0.026$. HW with \SI{1.5}{s} instead of \SI{1}{s} in its safety term lowers it from 28.2 to 21.7\,\% and
from 47.4 to 42.5\,\%, by $-6.5$ $[-9.3,-4.8]$ and $-4.9$ $[-7.9,-3.2]$, and its planner slows most before the attack
and loses most progress on the twins, $0.045$. Across the planners fine-tuned on generated scenarios, a lower
failure rate on the critical scenarios broadly goes with a lower progress on the benign twins, for both generators
(Table~\ref{tab:rq6}). The time headway of the label, not the generator, sets this trade-off. The constant is not
the same quantity in the two: the solver places the ego vehicle on the boundary of its clause, so the median headway
of our labels equals the constant, 1.4 to \SI{1.5}{s}, while the soft term of HW exceeds it, \SI{1.63}{s} at
\SI{1.5}{s}. The failure rate on the twins is also lower for the three \SI{1.5}{s} labels, 1.6 to 1.8\,\% against 5.4\,\% for
the replay, and on the 60 nuPlan logs the progress of every planner lies within 0.01 of the replay's and the failure
rate is unchanged. The specification adds a second constant. With the severity constant $a=4$ of
Section~\ref{sec:exp:severity} on the braking scenarios, the same clause lowers the failure rate by a further 2.5
$[1.4,4.1]$ and 3.4 $[1.9,6.0]$ percentage points, to 24.2 and 43.1\,\%, at a progress of $-0.042$; these labels
brake at \SI{4.0}{m/s^2} at the median against 2.9 for the default label and 2.0 for HW at \SI{1.5}{s}. It is the constant
that makes the test of Section~\ref{sec:exp:rq5} harder.

\section{DISCUSSION AND LIMITATIONS}\label{sec:discussion}

The ablations attribute the gain of the method to the representation of the roles in what is solved.
With the same clauses, step rule and budget, removing ownership leaves almost no valid scenarios, and
removing the stage order turns 41\,\% of the cut-ins into collisions with the unreacted ego vehicle. The step rule, in
contrast, can be exchanged for another without loss, and the multiplier schedule of the gate, given to
the hand-written sum, does not raise its rate. One
specification provides the objective, the order of the optimization and the acceptance test, and an LLM writes
it from a description.
Four limitations remain.

\emph{Drafting.} Most invalid attempts stem from specifications that the
solver cannot satisfy on the given window, such as a turning attacker confined to the ego's corridor.
The validity of a type varies strongly between windows and between drafts of the LLM. The valid attempts of
a type usually come from one of its windows, and four of the 11 generated types have a single valid attempt.
On CommonRoad, only three types gave valid attempts; for the others the attacker of the drafted specification left
the road or collided with a recorded vehicle. Pre-written clause templates for error-prone geometry and more windows per type may reduce this. We have
not evaluated them, and apart from the severity constant of Section~\ref{sec:exp:severity} and the headway clause of the
training labels (Section~\ref{sec:exp:rq6}) we did not vary the constants of the specification library
(Section~\ref{sec:method:language}). All experiments of Section~\ref{sec:experiments} have one attacker. A
specification with a further guided role compiles and is solved stage by stage without any change to the
solver. In a pilot, a second vehicle that drives beside the ego vehicle while the attacker cuts in was run
on the 10 nuScenes cut-in cases of Table~\ref{tab:rq2main} that have such a vehicle within \SI{35}{m}. One
case produced a scenario accepted by the monitor and the judges. The others failed because the blocker
could not reach the ego vehicle's side in time or because the cut-in itself failed on those cases. Because the lane clause of the cut-in specification forbids the ego vehicle's lane change anyway, the blocker did
not alter the response. The onset is still defined by the first role's observation clause,
and multi-role specifications were not evaluated further. Likewise, of the rows of Table~\ref{tab:catalogues}
marked expressible, the cut-out, the accelerating follower, the lane-changing ego vehicle, the motorcycle
scenarios and the ramp variants have specifications that the solver satisfies on synthetic scenes. They were not run on
recorded data before submission, because the case sets select one attacker on a through lane and hold no
motorcycles and no ramp or lane-changing ego windows.
When the clauses of one owner cannot be met together on a window, the solver returns no scenario and does not rank
the clauses. Minimum-violation planning under a priority order of STL specifications~\cite{halder2026lexicographic}
would instead return the scenario whose violations are smallest in the order of priority.

\emph{Data.} Signal states, reversing and parked attackers are not available
in the scenes we use, and junction and crossing scenarios are rare in them. The main experiments on
these families rest on a few dozen cases, and the training experiment covers cut-ins and lead-vehicle
braking on highways only. Each evaluated scenario has one attacker and one ego vehicle. The bystander
role and scenarios with several attackers are defined by the specification but not evaluated. All planners we test and train are learned planners that read trajectories. Sensor-level and end-to-end driving
systems, for which trajectory-level scenarios are the conditioning layer (Section~\ref{sec:intro}), are
not evaluated.

\emph{Seed on CommonRoad.} The CommonRoad seed is the plan of a diffusion-based motion planner, Diffusion
Planner~\cite{zheng2025diffusionplanner} trained on nuPlan. It collides, hits a recorded obstacle or
leaves the road on 37\,\% of the full set of windows and 54\,\% of the windows of the baseline comparison, and these
windows are dropped. The results do not depend on this seed. As an alternative, we used the plan of a
sampling-based planner~\cite{werling2010optimal}, which the admissibility condition drops on 5\,\% of the cases. The
two seeds give our method the same rate of valid scenarios on the same windows and attackers ($-0.02$
$[-0.08,+0.04]$ on 142 paired cases). With the sampling-planner seed, our method led HW by $+0.10$ $[+0.06,+0.15]$ pooled on
the harder windows this seed keeps, the cut-in collided with the unreacted ego vehicle in 46\,\% of the one-stage
attempts against 7\,\%, all 191 accepted scenarios were valid, and the learned planners failed $+7$ and $+18$
percentage points more often on our scenarios than on their seeds. The higher CommonRoad
EB rates in Table~\ref{tab:rq2main} reflect the case set, and so does the low failure rate of the planners on
the CommonRoad seeds in Table~\ref{tab:rq5}, which are Diffusion Planner's own plans on the windows where they
are valid. On CommonRoad, the slower diffusion ego leaves the
farther lead vehicles unreachable, and the dropped windows are mostly hand-crafted and SUMO maps, on which our method
is valid less often. As the edit is applied after sampling, the realism of
the edited trajectories rests on the dynamic bounds and on keeping the edit small and not on the diffusion
model. The attackers of all methods with a comparable rate of valid scenarios depart from logged driving about equally, in acceleration about twice as far as the
seed on nuScenes (Table~\ref{tab:realism}).

\emph{Training.} The time headway of the label sets a trade-off. With the default label, the fine-tuned planner
fails less than with the unedited seeds and keeps the progress of the replay-only planner on the benign twins, the unedited seeds of the critical scenarios, to
within about 0.01. With the \SI{1.5}{s} headway clause, and more so with the severity constant added, it fails less and
drives more cautiously: its progress on the benign twins falls by $0.037$ and $0.042$, about 3\,\%, as with HW
($0.026$ at its \SI{1}{s} headway, $0.045$ at \SI{1.5}{s}), while on the nuPlan logs it falls by 0.01 and the failure
rate there is unchanged. The same attacks with a different label change the result, and a higher headway in the
label of HW changes it alike, so the label shapes the trained planner as much as the attack does. Our ego vehicle reads the
attacker from the onset on, which the test requires; as a training label, it teaches the response and not the
caution before the attack that the safety term of HW teaches from the first step. A generator that serves training
has to choose the headway for that purpose, and we chose the clause's constant on a validation split.

\section{CONCLUSION}\label{sec:conclusion}

We presented a generator of safety-critical scenarios that states each scenario as a role-typed, causal
signal temporal logic specification. An LLM drafts the specification from a description and a type
checker accepts it. A compiler derives from it which agent each clause may move, the order in which the
attacker and the ego vehicle are solved, the onset that releases the response, and the monitor that
accepts the result. On nuScenes and CommonRoad, judged by independent judges that share no code with the
specification, one parameter set produced valid scenarios more often than weighted objectives tuned per family. All but one of the 204 scenarios the monitor accepted were valid, and the method produced at least one valid scenario for 11 of the 16 NHTSA
pre-crash types that the public data can express. Without ownership the attack was cancelled, and without the stage order 41\,\% of the cut-ins became collisions with
the unreacted ego vehicle. Solving the seeds of many windows in one batch raised the throughput about ninefold over one window at a time. The generated
scenarios made learned planners fail where failure was avoidable. Fine-tuning on them lowered the failure rate of a learned planner on critical scenarios, and the time headway set by one clause of the label set the size of the gain. Future work
includes clause templates for the LLM, the choice of the label's time headway for training, and scenarios that adapt the
attack to the planner under test.

\renewcommand{\dblfloatpagefraction}{0.45}\renewcommand{\dbltopfraction}{0.95}\renewcommand{\floatpagefraction}{0.45}
\appendices
\section{SPECIFICATIONS OF THE FOUR MAIN FAMILIES}\label{app:specs}

Tables~\ref{tab:spec:cutin} to~\ref{tab:spec:ltap} list the specifications of the four families of the main
experiments, with the default parameters with which all experiments ran. The formulas are those the compiler
evaluates, with the helper predicates of the library expanded. A negated predicate over a single agent is written with
the opposite comparison, which has the same robustness, and the negation of a predicate over a role with
several agents is existential (Section~\ref{sec:problem:specification}). All four specifications use
\begin{equation*}
\begin{aligned}
&\pred{rss\_margin}(i,j;L,a_i)=\pred{gap\_lon}(i,j;L)\\
&\quad-\max\Bigl(d_{\min},\ t_{\mathrm{r}} v_i+\frac{v_i^2}{2a_i}-\frac{v_j^2}{2a_j}\Bigr)
\end{aligned}
\end{equation*}
with the response time $t_{\mathrm{r}}=\delta+\SI{0.2}{s}$, the lead vehicle's braking $a_j=\SI{4.5}{m/s^2}$ and
$d_{\min}=\SI{2}{m}$. The predicate $\pred{box\_sep}^{\mathrm{rear}}(\ego,j)$ equals $\pred{box\_sep}(\ego,j)$
except at the steps at which agent $j$ is excused, where it counts as satisfied: before the first contact of the
pair, a step at which $j$ drives in the ego vehicle's direction, with headings less than \SI{90}{\degree}
apart, and its center lies behind the ego vehicle's rear bumper, and, from the first contact on, every step, if this
held at the first contact. All agents that the method does not change belong to the role $\oth$.

{\sloppy The LLMs of Section~\ref{sec:exp:rq1} were accessed through OpenRouter under the identifiers
\texttt{anthropic/\allowbreak claude-\allowbreak opus-\allowbreak 5.5}, \texttt{openai/\allowbreak gpt-\allowbreak 6.1-\allowbreak sol}, \texttt{deepseek/\allowbreak deepseek-\allowbreak v4-\allowbreak pro-\allowbreak 0813},
\texttt{anthropic/\allowbreak claude-\allowbreak sonnet-\allowbreak 5.5}, \texttt{openai/\allowbreak gpt-\allowbreak 6-\allowbreak luna} and \texttt{deepseek/\allowbreak deepseek-\allowbreak v4.1-\allowbreak flash},
each restricted to its first-party provider, with reasoning enabled, up to 32,768 output tokens and no sampling
parameter set; the coverage experiment used \texttt{deepseek/\allowbreak deepseek-\allowbreak v4-\allowbreak pro-\allowbreak 0813}.\par}

\begin{table*}[tp]
\caption{Cut-In Specification\label{tab:spec:cutin}}
\centering
\footnotesize
\setlength{\tabcolsep}{4pt}
\relpenalty=10000 \binoppenalty=0 %
\begin{tabular}{@{}llll>{\raggedright\arraybackslash}p{0.56\textwidth}@{}}
\toprule
Clause & Owner & Reads & Kind & Formula \\
\midrule
\cl{cut} & $\att$ & $\ego$ & attack &
$\Fop_{[1,4]}\bigl(\Gop_{[0,0.8]}\,|\pred{lat\_off}(\att;\mathrm{lane}(\ego))|<\SI{0.5}{m}
\wedge\pred{gap\_lon}(\ego,\att;\mathrm{lane}(\ego))<\SI{15}{m}
\wedge\pred{gap\_lon}(\ego,\att;\mathrm{lane}(\ego))>\SI{1}{m}\bigr)
\wedge\Fop_{[1,T]}\bigl(\pred{lat\_sep}(\ego,\att;\mathrm{lane}(\ego))<0
\wedge\pred{gap\_lon}(\ego,\att;\mathrm{lane}(\ego))>\SI{-0.5}{m}
\wedge\pred{rss\_margin}(\ego,\att;\mathrm{lane}(\ego),\SI{8}{m/s^2})<0\bigr)$ \\
\cl{avoidable} & $\att$ & $\ego$ & require &
$|\pred{lat\_off}(\att;\mathrm{lane}(\ego))|>\SI{1.9}{m}\;\Uop_{[0,T]}\;
\bigl(|\pred{lat\_off}(\att;\mathrm{lane}(\ego))|<\SI{1.9}{m}
\wedge\pred{rss\_margin}(\ego,\att;\mathrm{lane}(\ego),\SI{4.5}{m/s^2})>0\bigr)$ \\
\cl{no\_hit} & $\att$ & $\ego$ & require & $\Gop_{[0,T]}\,\pred{box\_sep}(\ego,\att)>\SI{0.05}{m}$ \\
\cl{att\_stays} & $\att$ & $\ego$ & require & $\Gop_{[4,T]}\,|\pred{lat\_off}(\att;\mathrm{lane}(\ego))|<\SI{0.5}{m}$ \\
\cl{att\_road} & $\att$ & $\emptyset$ & require & $\Gop_{[0,T]}\,\pred{road\_margin}(\att;\mathrm{road})>0$ \\
\cl{observe} & $\att$ & $\emptyset$ & observe &
$|\pred{lat\_shift}(\att;\mathrm{lane}(\att))|>\SI{0.5}{m}$, sustained for $\sigma=\SI{0.4}{s}$ \\
\cl{ego\_clear} & $\ego$ & $\att$ & require &
$\Gop_{[t^\star+\delta,T]}\bigl(\bigl(|\pred{lat\_off}(\att;\mathrm{lane}(\ego))|>\SI{1.9}{m}
\vee\pred{gap\_lon}(\ego,\att;\mathrm{lane}(\ego))>\SI{0.5}{m}\bigr)
\wedge\pred{box\_sep}(\ego,\att)>\SI{0.05}{m}\bigr)$ \\
\cl{ego\_recover} & $\ego$ & $\att$ & require &
$\Fop_{[t^\star+\delta,T]}\Gop_{[0,T]}\,\pred{rss\_margin}(\ego,\att;\mathrm{lane}(\ego),\SI{4.5}{m/s^2})>0$ \\
\cl{ego\_lane} & $\ego$ & $\emptyset$ & require & $\Gop_{[t^\star+\delta,T]}\,|\pred{lat\_off}(\ego;\mathrm{lane}(\ego))|<\SI{0.8}{m}$ \\
\cl{no\_third\_party\_hit} & $\att$ & $\oth$ & require & $\Gop_{[0,T]}\,\pred{box\_sep}(\att,\oth)>\SI{0.3}{m}$ \\
\cl{ego\_no\_third\_party\_hit} & $\ego$ & $\oth$ & require & $\Gop_{[t^\star+\delta,T]}\,\pred{box\_sep}^{\mathrm{rear}}(\ego,\oth)>\SI{0.3}{m}$ \\
\cl{ego\_road} & $\ego$ & $\emptyset$ & require & $\Gop_{[t^\star+\delta,T]}\,\pred{road\_margin}(\ego;\mathrm{road})>0$ \\
\bottomrule
\multicolumn{5}{@{}p{0.97\textwidth}@{}}{\rule{0pt}{2.5ex}Priority $(\att,\ego)$, context role $\oth$,
reaction delay $\delta=\SI{0.6}{s}$, horizon $T=\SI{5}{s}$; no hold time of the attacker; the ego vehicle keeps
its seed until $t^\star+\delta$. Defaults: gap to the unreacted ego vehicle after the cut between \SI{1}{m} and
\SI{15}{m}, observation threshold \SI{0.5}{m}, threat deceleration \SI{8}{m/s^2}, the friction limit, against
\SI{4.5}{m/s^2} for avoidability and recovery. Lanes: $\mathrm{lane}(\ego)$, $\mathrm{lane}(\att)$ in the
observation clause, and the road corridor $\mathrm{road}$ with the scene's width profile.}
\end{tabular}
\end{table*}

\begin{table*}[tp]
\caption{Lead-Vehicle Braking Specification\label{tab:spec:eb}}
\centering
\footnotesize
\setlength{\tabcolsep}{4pt}
\relpenalty=10000 \binoppenalty=0 %
\begin{tabular}{@{}llll>{\raggedright\arraybackslash}p{0.56\textwidth}@{}}
\toprule
Clause & Owner & Reads & Kind & Formula \\
\midrule
\cl{quiet} & $\att$ & $\emptyset$ & check & $\Gop_{[0,0.6]}\,\neg\Gop_{[0,0.3]}\,\pred{acc}(\att)<\SI{-2}{m/s^2}$ \\
\cl{brake} & $\att$ & $\ego$ & attack &
$\Fop_{[0.8,2]}\Gop_{[0,1]}\,\pred{acc}(\att)<\SI{-3.5}{m/s^2}
\wedge\Fop_{[2.4,T]}\bigl(\pred{lat\_sep}(\ego,\att;\mathrm{lane}(\ego))<0
\wedge\pred{gap\_lon}(\ego,\att;\mathrm{lane}(\ego))<\SI{-0.5}{m}
\wedge\pred{gap\_lon}(\att,\ego;\mathrm{lane}(\ego))<\SI{-0.5}{m}
\wedge\pred{box\_sep}(\ego,\att)<0\bigr)$ \\
\cl{att\_lane} & $\att$ & $\ego$ & require & $\Gop_{[0.8,T]}\,|\pred{lat\_off}(\att;\mathrm{lane}(\ego))|<\SI{0.8}{m}$ \\
\cl{att\_road} & $\att$ & $\emptyset$ & require & $\Gop_{[0.8,T]}\,\pred{road\_margin}(\att;\mathrm{road})>0$ \\
\cl{observe} & $\att$ & $\emptyset$ & observe & $\pred{acc}(\att)<\SI{-2}{m/s^2}$, sustained for $\sigma=\SI{0.4}{s}$ \\
\cl{ego\_clear} & $\ego$ & $\att$ & require &
$\Gop_{[t^\star+\delta,T]}\bigl(\pred{gap\_lon}(\ego,\att;\mathrm{lane}(\ego))>\SI{0.5}{m}\wedge\pred{box\_sep}(\ego,\att)>\SI{0.05}{m}\bigr)$ \\
\cl{ego\_lane} & $\ego$ & $\emptyset$ & require & $\Gop_{[t^\star+\delta,T]}\,|\pred{lat\_off}(\ego;\mathrm{lane}(\ego))|<\SI{0.8}{m}$ \\
\cl{no\_third\_party\_hit} & $\att$ & $\oth$ & require & $\Gop_{[0,T]}\,\pred{box\_sep}(\att,\oth)>\SI{0.3}{m}$ \\
\cl{ego\_no\_third\_party\_hit} & $\ego$ & $\oth$ & require & $\Gop_{[t^\star+\delta,T]}\,\pred{box\_sep}^{\mathrm{rear}}(\ego,\oth)>\SI{0.3}{m}$ \\
\cl{ego\_road} & $\ego$ & $\emptyset$ & require & $\Gop_{[t^\star+\delta,T]}\,\pred{road\_margin}(\ego;\mathrm{road})>0$ \\
\bottomrule
\multicolumn{5}{@{}p{0.97\textwidth}@{}}{\rule{0pt}{2.5ex}Priority $(\att,\ego)$, context role $\oth$,
reaction delay $\delta=\SI{0.6}{s}$, horizon $T=\SI{5}{s}$; the attacker keeps its seed until the hold time
\SI{0.8}{s}, and the ego vehicle until $t^\star+\delta$; the outer interval of $\cl{quiet}$ ends \SI{0.2}{s} before the hold time and
is checked without optimizing. Defaults: braking \SI{3.5}{m/s^2} for \SI{1}{s}, starting between the hold time
and \SI{2}{s}; contact not before \SI{2.4}{s}; observation threshold \SI{-2}{m/s^2}. The attacker's lane and
road clauses start at the later of the hold time and \SI{0.5}{s}. Lanes: $\mathrm{lane}(\ego)$ and the road
corridor $\mathrm{road}$ with the scene's width profile.}
\end{tabular}
\end{table*}

\begin{table*}[tp]
\caption{VRU Crossing Specification\label{tab:spec:vru}}
\centering
\footnotesize
\setlength{\tabcolsep}{4pt}
\relpenalty=10000 \binoppenalty=0 %
\begin{tabular}{@{}llll>{\raggedright\arraybackslash}p{0.56\textwidth}@{}}
\toprule
Clause & Owner & Reads & Kind & Formula \\
\midrule
\cl{cross} & $\att$ & $\ego$ & attack &
$\Fop_{[1,4.5]}\bigl(|\pred{lat\_off}(\att;\mathrm{lane}(\ego))|<\SI{0.5}{m}\wedge\pred{gap\_lon}(\ego,\att;\mathrm{lane}(\ego))>0\bigr)
\wedge\Fop_{[2,T]}\bigl(\pred{lat\_sep}(\ego,\att;\mathrm{lane}(\ego))<0
\wedge\pred{gap\_lon}(\ego,\att;\mathrm{lane}(\ego))<\SI{-0.5}{m}
\wedge\pred{gap\_lon}(\att,\ego;\mathrm{lane}(\ego))<\SI{-0.5}{m}
\wedge\pred{box\_sep}(\ego,\att)<0\bigr)$ \\
\cl{avoidable} & $\att$ & $\ego$ & require &
$|\pred{lat\_off}(\att;\mathrm{lane}(\ego))|>\SI{1.2}{m}\;\Uop_{[0,T]}\;
\bigl(|\pred{lat\_off}(\att;\mathrm{lane}(\ego))|<\SI{1.2}{m}
\wedge\pred{rss\_margin}(\ego,\att;\mathrm{lane}(\ego),\SI{4.5}{m/s^2})>0\bigr)$ \\
\cl{no\_early\_hit} & $\att$ & $\ego$ & require & $\Gop_{[0,1.8]}\,\pred{box\_sep}(\ego,\att)>\SI{0.05}{m}$ \\
\cl{walking\_speed} & $\att$ & $\emptyset$ & require & $\Gop_{[0,T]}\,\pred{speed}(\att)<\SI{3}{m/s}$ \\
\cl{observe} & $\att$ & $\emptyset$ & observe & $\pred{road\_margin}(\att;\mathrm{road})>0$, sustained for $\sigma=\SI{0.4}{s}$ \\
\cl{ego\_clear} & $\ego$ & $\att$ & require &
$\Gop_{[t^\star+\delta,T]}\bigl(\pred{gap\_lon}(\ego,\att;\mathrm{lane}(\ego))>\SI{0.5}{m}\wedge\pred{box\_sep}(\ego,\att)>\SI{0.05}{m}\bigr)$ \\
\cl{ego\_lane} & $\ego$ & $\emptyset$ & require & $\Gop_{[t^\star+\delta,T]}\,|\pred{lat\_off}(\ego;\mathrm{lane}(\ego))|<\SI{0.8}{m}$ \\
\cl{no\_third\_party\_hit} & $\att$ & $\oth$ & require & $\Gop_{[0,T]}\,\pred{box\_sep}(\att,\oth)>\SI{0.3}{m}$ \\
\cl{ego\_no\_third\_party\_hit} & $\ego$ & $\oth$ & require & $\Gop_{[t^\star+\delta,T]}\,\pred{box\_sep}^{\mathrm{rear}}(\ego,\oth)>\SI{0.3}{m}$ \\
\cl{ego\_road} & $\ego$ & $\emptyset$ & require & $\Gop_{[t^\star+\delta,T]}\,\pred{road\_margin}(\ego;\mathrm{road})>0$ \\
\bottomrule
\multicolumn{5}{@{}p{0.97\textwidth}@{}}{\rule{0pt}{2.5ex}The attacker is a pedestrian. Priority $(\att,\ego)$,
context role $\oth$, reaction delay $\delta=\SI{0.6}{s}$, horizon $T=\SI{5}{s}$; no hold time of the attacker;
the ego vehicle keeps its seed until $t^\star+\delta$. Defaults: contact not before \SI{2}{s}, no contact up to
\SI{0.2}{s} before it; walking speed below \SI{3}{m/s}; the lateral bound \SI{1.2}{m} of $\cl{avoidable}$ is
half the sum of the widths \SI{1.9}{m} and \SI{0.5}{m}. The pedestrian starts off the road and has no road
clause; its actions are bounded at \SI{2}{m/s^2}, \SI{3}{m/s^2} lateral acceleration, \SI{1.5}{rad/s} yaw rate
and \SI{3.5}{m/s}, without a curvature bound. Lanes: $\mathrm{lane}(\ego)$ and the road corridor
$\mathrm{road}$ with the scene's width profile.}
\end{tabular}
\end{table*}

\begin{table*}[tp]
\caption{Left-Turn-Across-Path Specification\label{tab:spec:ltap}}
\centering
\footnotesize
\setlength{\tabcolsep}{4pt}
\relpenalty=10000 \binoppenalty=0 %
\begin{tabular}{@{}llll>{\raggedright\arraybackslash}p{0.56\textwidth}@{}}
\toprule
Clause & Owner & Reads & Kind & Formula \\
\midrule
\cl{turn\_hit} & $\att$ & $\ego$ & attack & $\Fop_{[2,T]}\,\pred{box\_sep}(\ego,\att)<0$ \\
\cl{follow\_turn} & $\att$ & $\emptyset$ & require &
$\Gop_{[0.5,T]}\,|\pred{lat\_off}(\att;\mathrm{left\_turn\_of}(\mathrm{lane}(\att)))|<\SI{0.8}{m}$ \\
\cl{att\_road} & $\att$ & $\emptyset$ & require &
$\Gop_{[0.5,T]}\,\pred{road\_margin}(\att;\mathrm{left\_turn\_of}(\mathrm{lane}(\att)))>0$ \\
\cl{avoidable} & $\att$ & $\ego$ & require &
$|\pred{lat\_off}(\att;\mathrm{lane}(\ego))|>\SI{1.9}{m}\;\Uop_{[0,T]}\;
\bigl(|\pred{lat\_off}(\att;\mathrm{lane}(\ego))|<\SI{1.9}{m}
\wedge\pred{rss\_margin}(\ego,\att;\mathrm{lane}(\ego),\SI{4.5}{m/s^2})>0\bigr)$ \\
\cl{no\_early\_hit} & $\att$ & $\ego$ & require & $\Gop_{[0,1.8]}\,\pred{box\_sep}(\ego,\att)>\SI{0.05}{m}$ \\
\cl{observe} & $\att$ & $\emptyset$ & observe &
$|\pred{lat\_shift}(\att;\mathrm{lane}(\att))|>\SI{0.5}{m}$, sustained for $\sigma=\SI{0.4}{s}$ \\
\cl{ego\_clear} & $\ego$ & $\att$ & require &
$\Gop_{[t^\star+\delta,T]}\bigl(\pred{gap\_lon}(\ego,\att;\mathrm{lane}(\ego))>\SI{0.5}{m}\wedge\pred{box\_sep}(\ego,\att)>\SI{0.05}{m}\bigr)$ \\
\cl{ego\_lane} & $\ego$ & $\emptyset$ & require & $\Gop_{[t^\star+\delta,T]}\,|\pred{lat\_off}(\ego;\mathrm{lane}(\ego))|<\SI{0.8}{m}$ \\
\cl{no\_third\_party\_hit} & $\att$ & $\oth$ & require & $\Gop_{[0,T]}\,\pred{box\_sep}(\att,\oth)>\SI{0.3}{m}$ \\
\cl{ego\_no\_third\_party\_hit} & $\ego$ & $\oth$ & require & $\Gop_{[t^\star+\delta,T]}\,\pred{box\_sep}^{\mathrm{rear}}(\ego,\oth)>\SI{0.3}{m}$ \\
\cl{ego\_road} & $\ego$ & $\emptyset$ & require & $\Gop_{[t^\star+\delta,T]}\,\pred{road\_margin}(\ego;\mathrm{road})>0$ \\
\bottomrule
\multicolumn{5}{@{}p{0.97\textwidth}@{}}{\rule{0pt}{2.5ex}Priority $(\att,\ego)$, context role $\oth$,
reaction delay $\delta=\SI{0.6}{s}$, horizon $T=\SI{5}{s}$; no hold time of the attacker; the ego vehicle keeps
its seed until $t^\star+\delta$. Defaults: contact not before \SI{2}{s}, no contact up to \SI{0.2}{s} before it;
observation threshold \SI{0.5}{m}; the turn path and the attacker's road clause start at \SI{0.5}{s}. Lanes:
$\mathrm{lane}(\ego)$, $\mathrm{lane}(\att)$, the turn path $\mathrm{left\_turn\_of}(\mathrm{lane}(\att))$
from the junction connectivity of the map, replaced by $\mathrm{right\_turn\_of}(\mathrm{lane}(\att))$ on maps
with left-hand traffic, the road corridor around the turn path in $\cl{att\_road}$, and the road corridor
$\mathrm{road}$ with the scene's width profile.}
\end{tabular}
\end{table*}

\section{COVERAGE OF SCENARIO CATALOGUES}\label{app:coverage}
Table~\ref{tab:rq7} lists the 16 NHTSA pre-crash types that the agents and maps of the public data can express, with
the share of crashes of each type, the number of valid scenarios over the executed attempts per platform, and the number
of seed windows with at least one valid scenario (Section~\ref{sec:exp:rq7}).
\begin{table*}[tp]
\caption{NHTSA Pre-Crash Types Run End to End\label{tab:rq7}}
\centering\scriptsize\setlength{\tabcolsep}{5pt}\renewcommand{\arraystretch}{0.95} %
\begin{tabular}{@{}llrrrrr@{}}
\toprule
Type & Name & Freq.\ [\%] & Valid & nuSc. & CR & Win. \\
\midrule
N01 & Lead Vehicle Stopped & 16.4 & 4/16 & 0/8 & 4/8 & 1 \\
N03 & Vehicle(s) Turning at Non-Signalized Junctions & 7.3 & 4/24 & 4/16 & 0/8 & 2 \\
N04 & Lead Vehicle Decelerating & 7.2 & 1/16 & 1/8 & 0/8 & 1 \\
N06 & Vehicle(s) Changing Lanes -- Same Direction & 5.7 & 9/16 & 5/8 & 4/8 & 2 \\
N08 & Straight Crossing Paths at Non-Signalized Junctions & 4.4 & 0/24 & 0/16 & 0/8 & 0 \\
N11 & LTAP/OD at Signalized Junctions & 3.7 & 0/16 & 0/8 & 0/8 & 0 \\
N12 & Lead Vehicle Moving at Lower Constant Speed & 3.5 & 1/16 & 1/8 & 0/8 & 1 \\
N13 & LTAP/OD at Non-Signalized Junctions & 3.2 & 2/16 & 2/8 & 0/8 & 1 \\
N15 & Vehicle(s) Not Making a Maneuver -- Opposite Direction & 2.1 & 0/24 & 0/16 & 0/8 & 0 \\
N17 & Vehicle(s) Drifting -- Same Direction & 1.6 & 7/24 & 6/16 & 1/8 & 2 \\
N27 & Pedestrian Crash Without Prior Vehicle Maneuver & 0.7 & 4/16 & 4/16 & -- & 1 \\
N28 & Vehicle Turning Right at Signalized Junctions & 0.6 & 1/24 & 1/16 & 0/8 & 1 \\
N30 & Pedalcyclist Crash Without Prior Vehicle Maneuver & 0.4 & 1/16 & 1/16 & -- & 1 \\
N32 & Pedalcyclist Crash With Prior Vehicle Maneuver & 0.3 & 0/16 & 0/16 & -- & 0 \\
N34 & Lead Vehicle Accelerating & 0.3 & 0/16 & 0/8 & 0/8 & 0 \\
N35 & Vehicle(s) Making a Maneuver -- Opposite Direction & 0.3 & 4/24 & 4/16 & 0/8 & 1 \\
\bottomrule
\end{tabular}
\par\smallskip\parbox{\textwidth}{\footnotesize The 16 of the 37 types that the agents and maps of the public data can express, each run on the windows of both platforms where its agents exist, with DeepSeek V4-Pro, four descriptions and two drafts per window. Freq.: share of crashes~\cite{nhtsa2007typology}. Valid: valid attempts over executed attempts; nuSc., CR: the same per platform (--: no window). Win.: seed windows with at least one valid attempt.}
\end{table*}

Table~\ref{tab:catalogues} relates two further catalogues to the NHTSA types and to our runs: the traffic-related
critical scenarios of ISO 34502~\cite{iso34502}, Annex~B, which structures highway scenarios by road sector (main
roadway, merge zone, departure zone), the subject vehicle's behavior (lane keeping or lane change) and the motion of a
surrounding vehicle (cut-in, cut-out, acceleration, deceleration) into 24 general-vehicle scenarios, eight
motorcycle-specific ones and generic ones for other road users and obstructions; and the AEB test scenarios of the Euro
NCAP 2023 protocols for car-to-car and vulnerable road users. The perception- and vehicle-control-related scenarios of
ISO 34502 (Annexes~C and~D) and the lane-support scenarios of Euro NCAP concern the ego vehicle's sensing, control or
own lane departure and have no adversarial agent. They are therefore outside the scope of trajectory-level generation. For each row, the status is one of three: \emph{generated}, valid scenarios exist on the recorded data of
Section~\ref{sec:exp:rq7} or of the main experiments; \emph{expressible}, a specification in our language exists,
type-checks, and is solved by the fixed solver on a synthetic scene with the monitor and the independent judge
accepting the result, but no valid scenario on recorded data exists yet, for the reason given in the last column; and
\emph{out of scope}, which the public data or the trajectory-level setting does not provide.

\begin{table*}[tp]
\caption{Scenario Catalogues Against Our Runs\label{tab:catalogues}}
\centering\scriptsize\setlength{\tabcolsep}{3pt}\renewcommand{\arraystretch}{0.9}
\begin{tabular}{@{}>{\raggedright\arraybackslash}p{0.13\textwidth}>{\raggedright\arraybackslash}p{0.29\textwidth}>{\raggedright\arraybackslash}p{0.11\textwidth}l>{\raggedright\arraybackslash}p{0.35\textwidth}@{}}
\toprule
Catalogue & Scenario & NHTSA type & Status & Evidence, or what is missing \\
\midrule
\multicolumn{5}{@{}l}{\emph{ISO 34502 Annex B, general vehicle (Fig.\ B.6; the ego keeps its lane unless stated)}} \\
No.\ 1 & cut-in on the main roadway & N06 & generated & cut-in family (Table~\ref{tab:rq2main}), 9/16 in Table~\ref{tab:rq7}; Fig.~\ref{fig:gallery}(c) \\
No.\ 9, 17 & cut-in in the merge and departure zones & N06 & expressible & the same specification; no ramp windows in the case sets (NGSIM US-101 has them) \\
No.\ 2, 10, 18 & cut-out: the lead vehicle leaves the lane and reveals a slower vehicle & N01, N04 (the revealed lead vehicle) & expressible & specification \texttt{cutout}; needs windows with a slower vehicle ahead of the lead vehicle \\
No.\ 3, 11, 19 & a following vehicle accelerates into the ego & none & expressible & specification \texttt{rear\_approach} (the ego moves clear); the judges excuse the ego from a rear impact, so this threat needs its own judge item \\
No.\ 4, 12, 20 & the lead vehicle decelerates or stops & N01, N04, N12 & generated & lead-vehicle braking family (Table~\ref{tab:rq2main}); 4/16, 1/16, 1/16 in Table~\ref{tab:rq7} \\
No.\ 5--8, 13--16, 21--24 & the four motions while the ego changes lane & Following vehicle making a maneuver & expressible & specification \texttt{cutin\_lanechange} (the ego may settle in either lane); needs seed windows in which the ego changes lane \\
\multicolumn{5}{@{}l}{\emph{ISO 34502 Annex B, motorcycles and other road users}} \\
B.3, No.\ 1--8 & a motorcycle moves into or out of the in-lane position beside the ego & none & expressible & specification \texttt{moto\_filter}, lane sharing followed by a cut-in, solved with thin margins; no motorcycles in the case sets \\
B.4 & pedestrian or cyclist in the lane, as a simplification of No.\ 4 or 8 & N27, N30 & generated & VRU crossing family; 4/16 and 1/16 in Table~\ref{tab:rq7} \\
B.4 & obstruction in the lane & none & out of scope & a standing object is a property of the seed, not an attack: the type check rejects an attack clause the seed already meets \\
\midrule
\multicolumn{5}{@{}l}{\emph{Euro NCAP 2023 AEB car-to-car}} \\
CCRs & rear, stationary lead vehicle & N01 & generated & 4/16 in Table~\ref{tab:rq7}; Fig.~\ref{fig:gallery}(f) \\
CCRm & rear, lead vehicle at lower constant speed & N12 & generated & 1/16 in Table~\ref{tab:rq7} \\
CCRb & rear, braking lead vehicle & N04 & generated & lead-vehicle braking family (Table~\ref{tab:rq2main}) \\
CCFtap & the ego turns across an oncoming vehicle & N13, N11, N35 & out of scope & the LTAP family turns the attacker across the ego vehicle (N13, N35; Fig.~\ref{fig:gallery}(d)); in CCFtap the ego vehicle turns, and no window of the public data with a turning ego vehicle has an admissible seed \\
CCCscp & straight crossing paths at a junction & N08 & expressible & 24 attempts with LLM-drafted specifications, none valid: the drafted road clauses confine the crossing attacker to the ego's road corridor \\
CCFhos & oncoming vehicle drifts into the ego's lane & N15 & expressible & 24 attempts, none valid (same cause); specification \texttt{oncoming\_drift} \\
CCFhol & oncoming vehicle changes lane into the ego's lane & N35 & generated & 4/24 in Table~\ref{tab:rq7} \\
\multicolumn{5}{@{}l}{\emph{Euro NCAP 2023 AEB vulnerable road users}} \\
CPFA, CPNA, CPNCO & pedestrian crossing from the far or near side, also from behind an obstruction & N27 & generated & VRU crossing family; 4/16 in Table~\ref{tab:rq7}; Fig.~\ref{fig:gallery}(e) \\
CPLA & pedestrian ahead, walking in the ego's direction & N27 & expressible & specification \texttt{vru\_ahead} \\
CPTAns/no/fs/fo & the ego turns into a crossing pedestrian & Pedestrian crash with prior vehicle maneuver & out of scope & no window of the public data with a turning ego vehicle has an admissible seed \\
CPRs, CPRm & reversing into a pedestrian & Backing up & out of scope & no reverse motion in the scenes \\
CBFA, CBNA, CBNAO & cyclist crossing from the far or near side, also obstructed & N30 & generated & 1/16 in Table~\ref{tab:rq7} \\
CBLA & cyclist ahead, riding in the ego's direction & N30 & expressible & specification \texttt{vru\_ahead} with cyclist dynamics \\
CBTAn, CBTAf & the ego turns into a crossing cyclist & N32 & expressible & 16 attempts, none valid (turning ego windows) \\
CMRs, CMRb, CMFtap & motorcycle versions of CCRs, CCRb, CCFtap & N01, N04, N13 & expressible & the solver satisfies the lead-vehicle braking and LTAP specifications with a motorcycle attacker on the synthetic scenes; no motorcycles in the case sets \\
LSS, CM oncoming / overtaking & the ego itself drifts or overtakes & none & out of scope & no adversarial agent \\
\bottomrule
\end{tabular}
\par\smallskip\parbox{\textwidth}{\scriptsize Generated: valid scenarios on recorded data (counts of Table~\ref{tab:rq7}: valid over executed attempts). Expressible: a specification that the solver satisfies on a synthetic scene, no valid scenario on recorded data yet. ISO 34502 numbers: Fig.\ B.6 and B.9 of the standard; Euro NCAP abbreviations: the 2023 protocols.}
\end{table*}

\clearpage %

\bibliographystyle{IEEEtran}
\bibliography{references}

\end{document}